%% file: RL_iclr24_revise.tex
\documentclass{article} 
\usepackage{iclr2024_conference,times}

\input{math_commands.tex}

\input{myPKG.tex}

\input{new_front.tex}

\usepackage{hyperref}
\usepackage{url}

\title{Fast Value Tracking for Deep Reinforcement Learning}

\author{Frank Shih\\
Department of Statistics\\
Purdue University\\
West Lafayette, IN 47907, USA \\
\texttt{shih37@purdue.edu} \\
\And
Faming Liang  \\
Department of Statistics \\
Purdue University \\
West Lafayette, IN 47907, USA \\
\texttt{fmliang@purdue.edu} \\
}

\begin{document}

\maketitle

\begin{abstract}
   Reinforcement learning (RL) tackles sequential decision-making problems by creating
agents that interacts with their environment. However, existing algorithms often view these problem as 
static, focusing on point estimates for model parameters to maximize expected rewards, neglecting the stochastic dynamics of agent-environment interactions and the critical role of uncertainty quantification.
Our research leverages the Kalman filtering paradigm to introduce a novel and scalable sampling algorithm called Langevinized Kalman Temporal-Difference (LKTD) for deep reinforcement learning. This algorithm, grounded in Stochastic Gradient Markov Chain Monte Carlo (SGMCMC), efficiently draws samples from the posterior distribution of deep neural network parameters. Under mild conditions, we prove that the posterior samples generated by the LKTD algorithm converge to a stationary distribution. This convergence not only enables us to quantify uncertainties associated with the value function and model parameters but also allows us to monitor these uncertainties during policy updates throughout the training phase. The LKTD algorithm paves the way for more robust and adaptable reinforcement learning approaches.
\end{abstract}

\vspace{-0.1in}
\section{Introduction}
\vspace{-0.05in}

Over the last decade, RL has achieved remarkable successes across a diverse array of tasks, including robotics \citep{Kormushev2013ReinforcementLI}, video games \citep{Silver2016MasteringTG},  bidding strategies \citep{Jin2018RealTimeBW}, and ridesharing optimization \citep{Xu2018LargeScaleOD}. 
As a mathematical model, RL solves sequential decision-making problems by designing an agent that interacts with the environment, the goal is to learn an optimal policy that maximizes the expected total reward for the agent. 
Prominent value-based algorithms, including Temporal-difference (TD) learning \citep{Sutton:1988}, State–action–reward–state–action (SARSA) \citep{Sutton1998}, and Q-learning, aim to derive an optimal policy through learning values of states (or Q-values). Traditionally, these methods treat the state value (or Q-value) as a deterministic function, focusing on calculating point estimates of model parameters, thereby overlooking the inherent stochasticity in agent-environment interactions.

In the context of RL, a fair algorithm should exhibit the features: 
(i) {\it Uncertainty quantification}, which addresses the stochastic nature of the agent-environment interactions, thereby enhancing the robustness of the learned policy;  
(ii) {\it Dynamicity}, which considers the dynamics of the agent-environment interaction system, thereby enhancing the practicality of the RL technique;
(iii) {\it Nonlinear approximation}, which employs, for example, a deep neural network to approximate the value function,
thereby broadening the algorithm's applicability;   
(iv) {\it Computational efficiency}, which is scalable with respect to the model dimension
 and training sample size, facilitating online learning. 
Therefore, in RL, it is more suitable to treat values or model parameters as random variables rather than fixed unknowns, focusing on tracking dynamic changes rather than achieving point convergence during the policy learning process.

To achieve these goals, the Kalman Temporal Difference (KTD) framework has been studied for RL in the literature, as seen in references such as e.g., \cite{Geist2010ACM}, \cite{Tripp2013ApproximateKF}, and \cite{Shashua2020KalmanMB}. In these studies, values or their parameters are treated as random variables, and the focus is on the tracking property of the policy learning process. Specifically, KTD conceptualizes RL as a state-space model:
\begin{equation} \label{statespaceeq}
 \begin{split} 
  \theta_t &= \theta_{t-1}+w_t, \\ 
  \br_t &= h(\bx_t,\theta_t)+\eta_t, 
 \end{split}
\end{equation}  
where $\theta_t\in \mathbb{R}^p$ denotes the parameters at time step $t$ with dimension $p$,  $w_t \in \mathbb{R}^p$ 
and $\eta_t \in \mathbb{R}^n$ denote two independent 
multivariate Gaussian vectors, $\bx_t$ denotes 
a set of states and actions collected at time step $t$, 
$\br_t \in \mathbb{R}^n$ denotes a vector of rewards, 
$n$ denotes the number of samples, and $h(\cdot)$ is a function to be defined in Section \ref{KTDsection}. Within the framework of state-space models, the top equation in (\ref{statespaceeq}) is reffered to as the state evolution equation, while the bottom equation is known as the measurement equation. 
 Under the normality assumption and for a linear measurement equation, where 
 $h(\bx,\theta)$ is a linear function of $\theta$, 
 the Kalman filter \citep{Kalman1960ANA} is able to iteratively update the mean and variance 
 estimates for $\theta_t$ conditioned on the rewards $(\br_t,\br_{t-1},\ldots,\br_1)$,  enabling a proper quantification of the uncertainty associated with 
 the dynamic agent-environment interaction system. 
 However, when $h(\bx,\theta)$ becomes nonlinear, it necessitates the use of linearization techniques.
 Specifically, \cite{Geist2010ACM} employs Unscented Kalman Filter (UKF) \citep{UKF2000}, while \cite{Shashua2020KalmanMB} utilize Extended Kalman Filter (EKF) \cite{EKF1979OptimalF} to approximate the covariance matrices of $\theta_t$. 
 Unfortunately, both UKF and EKF becomes computationally inefficient for high-dimensional parameter spaces, a common scenario when employing large-scale neural networks to approximate  $h(\cdot,\cdot)$. These filters require $O(p^2)$ additional space to store the covariance matrix and $O(np^2)$ for matrix multiplications at each iteration. 
 Moreover, the linearization operation involved in these algorithms can degrade the accuracy of estimation. 
 To address the limitations encountered by KTD, we reformulate RL as the following state space model: 
\begin{equation} \label{statespaceeq2}
 \begin{split} 
\theta_t &= \theta_{t-1}+ \frac{\epsilon_t}{2} \nabla_{\theta} \log \pi(\theta_{t-1})
 + w_t, \\ 
\br_t &= h(\bx_t,\theta_t)+\eta_t, 
 \end{split}
\end{equation}  
where $w_t \sim N(0, \epsilon_t I_p)$, 
$\pi(\theta)$ represents a prior density function we impose on $\theta$, and $\{\epsilon_t: t=1,2,\ldots\}$ is a positive sequence decaying to zero. 
Additionally, we propose to update $\theta_t$ using the Langevinized 
Ensemble Kalman Filter (LEnKF) algorithm \citep{Zhang2021LEnKF}.  
With the formulation (\ref{statespaceeq2}) and the LEnKF algorithm, 
we show in  Section \ref{LKTDtheory} that $\theta_t$ converges to a 
proper distribution as the learning horizon $t \to \infty$, enabling the uncertainty 
associated with the dynamic agent-environment interaction system to be properly 
quantified. Inclusion of the prior information in the state evolution equation 
generally robustifies the performance of the RL algorithm. 
For instance, when employing a large-scale deep neural network to approximate the function $h(\cdot,\cdot)$, selecting an appropriate $\pi(\cdot)$, such as a mixture Gaussian distribution, 
can lead to the sparsification of the neural network. This enhances the robustness of the learned policy according to the theory of sparse deep learning \citep{SunSLiang2022sparseDNN}.  
Compared to existing KTD algorithms, the proposed algorithm can directly handle a nonlinear function $h(\cdot,\cdot)$ without the need for a linearization operator.
The proposed algorithm enables fast value tracking at a complexity of 
$O(np)$ per iteration, scalable for large-scale neural networks. It also enhances memory-efficiency as it replaces the storage for the covariance matrix with particles, representing samples of $\theta$. It is worth noting that both the sample size $n$ and the number of particles retained during the algorithm's execution are typically significantly smaller than the parameter size $p$. Lastly, we extend the convergence theory of our proposed algorithm to include scenarios that utilize replay buffers, thereby expanding its applicability beyond the on-policy framework.

Compared to model (\ref{statespaceeq}), our new formulation imposes slightly more restrictions on the variability of $\theta$ through the prior $\pi(\theta)$, while still accounting for the dynamics of the system. However, these restrictions do not diminish the generality and adaptivity of the model (\ref{statespaceeq2}), thanks to the universal approximation ability of deep neural networks that will be used in approximating $h(\cdot,\cdot)$ in this paper.

\paragraph{Related Works} Bootstrapped DQN \citep{BootDQN} and Quantile Regression DQN \citep{bellemare2017distributional} also aim to learn the uncertainty estimates for the value function, but they are not formulated under the KTD framework. 

\vspace{-0.1in}
\section{Background} \label{backsection}

\vspace{-0.1in}
\subsection{Markov Decision Process}

The standard RL procedure aims to learn an optimal policy from the interaction experiences between an agent and an environment, where the optimal policy maximizes the agent's expected total reward. The RL procedure can be described by a Markov Decision Process (MDP) represented by $\{\mathcal{S}, \mathcal{A}, \mathcal{P}, r, \gamma\}$, where $\mathcal{S}$ is set of states, $\mathcal{A}$ is a finite set of actions, $\mathcal{P}:\mathcal{S}\times\mathcal{A}\times\mathcal{S}\rightarrow \mathbb{R}$ is the state transition probability from state $s$ to state $s'$ by taking action $a$, denoted by $\mathcal{P}(s'| s,a)$, $r(s,a)$ is a random reward received from taking action $a$ at state $s$, and $\gamma\in (0,1)$ is a discount factor.  At each time step $t$, the agent observes state $s_{t}\in \mathcal{S}$ and takes action $a_t\in\mathcal{A}$ according to policy $\rho$ with probability $P_{\rho}(a|s)$, 
then the environment returns a reward $r_t = r(s_t,a_t)$ and a new state $s_{t+1}\in\mathcal{S}$. For a given policy $\rho $, the performance is measured by the state value function $V_\rho (s) = \E_\rho[\sum_{t=0}^\infty \gamma^t r_t| s_0=s]$ and the state-action value function $Q_\rho (s,a) = \E_\rho[\sum_{t=0}^\infty \gamma^t r_t| s_0=s, a_0=a]$, 
which are called $V$-function and $Q$-function, respectively. 
  Both functions satisfy the Bellman equation:
\begin{equation} \label{bellmaneq}
    \begin{split}
        V_\rho(s) &= \E_\rho [r(s,a) + \gamma V_\rho(s')], \\
        Q_\rho(s,a) &= \E_\rho [r(s,a) + \gamma Q_\rho(s',a')], 
    \end{split}
\end{equation}
where $s'\sim \mathcal{P}(\cdot|s,a)$, $a\sim P_\rho(\cdot | s)$, $a'\sim P_\rho(\cdot | s')$,
and the expectation $\E_{\rho}[\cdot]$ is taken over the transition probability distribution $\mathcal{P}$ for a given policy $\rho$.

\vspace{-0.05in}
\subsection{Kalman Temporal Difference (KTD) Algorithms} \label{KTDsection}
\vspace{-0.05in}

Let $\bs_t=(s_t^{(1)},s_t^{(2)},\ldots,s_t^{(n)})^T$, 
$\ba_t=(a_t^{(1)}, a_t^{(2)}, \ldots, a_t^{(n)})^T$,
and $\br_t=(r_t^{(1)}, r_t^{(2)}, 
\ldots, r_t^{(n)})^T$ denote, respectively, a vector of $n$ states, actions, rewards  collected at time step $t$.  
Given the Bellman equation, 
the function $h(\bx_t,\theta_t)$ in (\ref{statespaceeq}) can be expressed as 
\begin{equation} \label{h-function}
h(\bx_t,\theta_t)= \begin{cases} 
 V_{\theta_t}(\bs_t)-\gamma V_{\theta_t}(\bs_{t+1}), & \mbox{for $V$-function}, \\
 Q_{\theta_t}(\bs_t,\ba_t)-\gamma Q_{\theta_t}(\bs_{t+1},\ba_{t+1}),  & \mbox{ for $Q$-function}, \\
 \end{cases}
\end{equation} 
where $\bx_t=\{\bs_t,\ba_t,\bs_{t+1},\ba_{t+1}\}$, 
$V_{\theta_t}(\bs_t):=(V_{\theta_t}(s_t^{(1)}), V_{\theta_t}(s_t^{(2)}), \ldots, V_{\theta_t}(s_t^{(n)}))^T$, and $Q_{\theta_t}(\bs_t,\ba_t):=( Q_{\theta_t}(s_t^{(1)},a_t^{(1)})$, 
 $Q_{\theta_t}(s_t^{(2)},a_t^{(2)}), \ldots, Q_{\theta_t}(s_t^{(n)},a_t^{(n)}))^T$. 
\textcolor{black}{In this paper, we focus on Q-functions only, however, our algorithm also works for V-functions as indicated below.}
 The KTD framework works under the Gaussian assumption, i.e., $\theta_t$ follows a Gaussian 
 distribution at each stage $t=1,2,\ldots$.  To address the nonlinearity of the function 
 $h(\bx,\theta)$, the KOVA algorithm, as proposed by \cite{Shashua2020KalmanMB}, employs the Extended Kalman Filter (EKF) technique for calculating the mean and covariance matrices of $\theta_t$. This approach involves linearizing
 $h(\bx,\theta)$ based on the first-order Taylor expansion, namely: 
 \[
 h(\bx_{t},\theta) \approx h(\bx_{t},\hat{\mu}_{t-1})+\nabla_{\theta} h(\bx_{t},\hat{\mu}_{t-1})^T (\theta-\hat{\mu}_{t-1}),
 \]
 where $\hat{\mu}_{t-1}$ denotes the estimator for the mean of $\theta_{t-1}$.   
 The KOVA algorithm, detailed in Algorithm \ref{alg:KOVA}, however, encounters several significant challenges:  (i) the approximation accuracy for the true filtering distribution of $\theta_t$ is unknown; (ii) it exhibits high computational complexity $O(np^2)$; and (iii)  it demands considerable memory complexity, necessitating $O(p^2)$ additional space for the covariance matrix.
Alternatively, \cite{Geist2010ACM} recommended the implementation of KTD using Unscented Kalman Filter (UKF). Nonetheless, this alternative algorithm encounters similar challenges to those faced by KOVA, including issues related to approximation accuracy, computational complexity, and memory requirements.

\vspace{-0.05in}
\section{Langevinized Kalman Temporal Difference Algorithm} \label{LKTDtheory}
\vspace{-0.05in}
To overcome the limitations encountered by existing KTD algorithms, we introduce an approach that integrates KTD with the LEnKF algorithm, leading to the development of the Langevinized Kalman Temporal Difference (LKTD) algorithm.  
The LEnKF is a reformulation of the Ensemble Kalman filter (EnKF) \citep{Evensen1994} under the framework of Langevin dynamics.  
The LEnKF inherits the forecast-analysis procedure from the EnKF and the use of
minibatch data from the Stochastic Gradient Langevin Dynamics (SGLD) algorithm \citep{Welling2011BayesianLV}, making it scalable with respect to the state dimension $p$ and the mini-batch size $n$.  
Distinctively, the LEnKF algorithm is designed to converge to the accurate filtering distribution, setting it apart from the traditional EnKF.

\vspace{-0.05in}
\subsection{The LKTD Algorithm}
\vspace{-0.05in}

The LKTD algorithm is designed to solve the RL problem by framing it within the state-space model outlined in equation (\ref{statespaceeq2}). As previously explained, this model differs from the model (\ref{statespaceeq}) by incorporating the prior information, $\pi(\theta)$, into the state evolution equation.
This incorporation generally enhances the robustness of the algorithm, especially when using a deep neural network to approximate the function $h(\bx,\theta)$. 
Next, we can apply the variance splitting technique \citep{Zhang2021LEnKF} to convert the model (\ref{statespaceeq}) into a state-space model with a linear measurement equation, while allowing the state evolution equation to be nonlinear. The variance splitting technique can be described as follows. 

Without loss of generality, let's assume that $\eta_t \sim N(0,\sigma^2I_n)$ for each stage $t$, where $I_n$ is an $n\times n$-identity matrix.  By the state augmentation approach, we define 
\begin{equation} \label{hiereq}
    \varphi_t = 
    \begin{pmatrix}
    \theta_t\\
    \xi_t
    \end{pmatrix}
    , \quad \xi_t = h(\vx_t;\theta_t) + u_t, \quad u_t \sim N(0, \alpha \sigma^2 I_n),
\end{equation}
where $\xi_t$ is an $n$-dimensional vector, and $0<\alpha<1$ is a pre-specified constant. Suppose that $\theta_t$ has a prior distribution $\pi(\theta)$ as specified previously,  the joint density function of $\varphi_t = (\theta_t^\top, \xi_t^\top)^\top$ is given by  $\pi(\varphi_t) = \pi(\theta_t)\pi(\xi_t|\theta_t)$, where $\xi_t|\theta_t\sim N(h(\vx_t;\theta_t), \alpha \sigma^2 I)$.
Based on Langevin dynamics, we can reformulate (\ref{statespaceeq2}) as the following 
model: 
\begin{equation}\label{eq:nonlinear state-space model}
\begin{split}
\varphi_t &= \varphi_{t-1} + \frac{\epsilon_t}{2} \frac{n}{\mathcal{N}} \nabla_{\varphi} \log \pi(\varphi_{t-1}) + \tilde{w}_t,\\
\vr_t &= H_t \varphi_t + v_t,
\end{split}
\end{equation}
where $\mathcal{N}>0$, $\tilde{w}_t\sim N(0, \frac{n}{\mathcal{N}} B_t)$, $B_t = \epsilon_t I_{\tilde{p}}$, $\tilde{p}=p+n$ is the dimension of $\varphi_t$; $H_t = ({\bf 0},I_n)$ such that $H_t \varphi_t = \xi_t$; $v_t\sim N(0, (1-\alpha)\sigma^2 I_n)$, which is independent of $\tilde{w}_t$ for all $t$.  We call $\mathcal{N}$ the pseudo-population size, which scales 
uncertainty of the estimator of the system. Refer to Lemma \ref{lem:1} 
and Theorem \ref{thm:SGLD} for mathematical justifications for this issue.

By (\ref{hiereq}) and (\ref{eq:nonlinear state-space model}),  
$\vx_t$, $\btheta_t$, $\xi_t$ and $\vr_t$ form a hierarchical model with the conditional distribution
\begin{equation} \label{hiereq2}
\xi_t|\vr_t, \vx_t,\btheta_t \sim \mathcal{N}(\alpha \vr_t+(1-\alpha) h(\vx_t;\btheta_t), \alpha(1-\alpha) \sigma^2 I_n),
\end{equation}
which, as shown by \cite{Zhang2021LEnKF}, eventually leads to an efficient particle filtering  algorithm for handling the state-space models with a nonlinear measurement equation. Similar to LEnKF, we adopt the forecast-analysis procedure from EnKF to the model (\ref{eq:nonlinear state-space model}), leading to Algorithm \ref{alg:LKTD}. It works in a single chain, different from particle filtering algorithms.  
The time complexity of the algorithm  is $O(n p)$. This attractive time complexity is due to the special structure
of $H_t$, rendering the matrix $K_t$ and equation \ref{eq: analysis} easily computed. 
Regarding the settings of $\mathcal{K}$ and $\alpha$, we make the following remarks: First, as indicated by  
 the Kalman gain matrix $K_{t,k}$, only the $\xi$-component of $\varphi_{t,k}$ is updated 
  at each analysis step. Generally, $\xi_{t,k}$ converges rapidly, benefiting from second-order gradient information.  Thus, $\mathcal{K}$ does not need to be excessively large.  
  Based on the property (\ref{hiereq2}),  \cite{Zhang2021LEnKF} demonstrated that LEnKF acts as a variance reduction version of SGLD if $0.5 < \alpha<1$, recommending $\alpha$ be set close to 1.   
 In this paper, we default $\mathcal{K}=5$ and $\alpha=0.9$, initializing $\xi_{t,0}$ by $\br_t$ at each time step $t$ to enhance the convergence of the simulation.

\begin{algorithm}[htbp]
\SetAlgoLined
\textbf{Initialization:} Draw $\theta_0^a\in \mathbb{R}^p$ drawn from the prior distribution $\pi(\theta)$.

\For{t=1,2,\dots, T}{
    \textbf{Sampling:} With policy $\rho_{\theta_{t-1}^a}$, generate a set of $n$ transition tuples, 
    denoted by  $\vz_{t} = (\vr_t, \vx_t) :=  \{r_t^{(j)}, x_{t}^{(j)}\}_{j=1}^n$, where 
  $x_t^{(j)} = (s_t^{(j)}, a_t^{(j)}, s_{t+1}^{(j)}, a_{t+1}^{(j)})^T$  
   and $x_t^{(j)} = (s_t^{(j)}, a_t^{(j)}, s_{t+1}^{(j)})^T$ correspond to the 
   choices of the $Q$-function and  $V$-function in (\ref{h-function}), respectively.

 \For{k=1,2,\ldots,$\mathcal{K}$}{
      
       \textbf{Presetting:} Set $B_{t,k} = \epsilon_{t,k} I_{\tilde{p}}$, $R_t = 2(1-\alpha) \sigma^2 I$, and the Kalman gain matrix $K_{t,k} = B_{t,k} H_{t}^\top (H_{t} B_{t,k} H_{t}^\top + R_t)^{-1}$.

    \textbf{Forecast:} Draw $\tilde{w}_{t,k}\sim N_p(0, \frac{n}{\mathcal{N}} B_{t,k})$ and calculate
        \begin{equation}
            \varphi_{t,k}^{f} = \varphi_{t,k-1}^{a} + \frac{\epsilon_{t,k}}{2} \frac{n}{\mathcal{N}} \nabla_{\varphi} \log \pi (\varphi_{t,k-1}^{a}) + \tilde{w}_{t,k},
        \end{equation}
    where $\varphi_{t,0}^a =( {\theta_{t-1,\mathcal{K}}^a}^\top,  \vr_t^\top)^\top$ if $k=1$, and the gradient term is given by 
    \begin{equation}\label{eq:nonlinear gradient}
        \nabla_{\varphi} \log \pi (\varphi_{t,k-1}^{a}) = \begin{pmatrix}
        \nabla_\theta \log \pi(\theta_{t,k-1}) + \frac{1}{\alpha\sigma^2} 
        \frac{\mathcal{N}}{n} \nabla_\theta h(\vx_t; \theta_{t,k-1})(\xi_{t,k-1}- h(\vx_t; \theta_{t,k-1})) \\
        -\frac{1}{\alpha \sigma^2}  (\xi_{t,k-1}-h(\vx_t; \theta_{t,k-1}))
        \end{pmatrix}.
    \end{equation}
    
    \textbf{Analysis:} Draw $v_{t,k}\sim N_n(0, \frac{n}{\mathcal{N}} R_t)$ and calculate
        \begin{equation}\label{eq: analysis}
            \varphi_{t,k}^{a} = \varphi_{t,k}^{f} +K_{t,k}(\vr_{t} - H_{t} \varphi_{t,k}^{f} - v_{t,k}) = \varphi_{t,k}^{f} +K_{t,k}(\vr_{t} - \vr_{t,k}^{f }).
        \end{equation}
} 
}
\caption{Langevinized Kalman Temporal-Difference (LKTD)}
\label{alg:LKTD}
\end{algorithm}

\vspace{-0.05in}
\subsection{Convergence Theory} \label{theorysection}
\vspace{-0.05in}

To study the convergence of Algorithm \ref{alg:LKTD}, it suffices to study the convergence of Algorithm \ref{alg:prototype}, which ignores the inner iterations for imputing the latent variables $\xi_t$ and serves as the prototype of Algorithm \ref{alg:LKTD}.  
Lemma \ref{lem:1} shows that Algorithm \ref{alg:prototype} is actually an accelerated 
preconditioned SGLD algorithm. Its proof follows Theorem S1 of \cite{Zhang2021LEnKF} . 

\begin{lemma}
\label{lem:1}
Algorithm \ref{alg:prototype} implements a preconditioned SGLD algorithm, for which 
\begin{equation} \label{precondSGLD}
    \varphi_t^a = \varphi_{t-1}^a + \frac{\epsilon_t}{2}\Sigma_t \nabla_{\varphi} \log \pi(\varphi_{t-1}^a|\vz_{t}) + e_t,
\end{equation}
where $\vz_{t} = (\vr_t, \vx_t)$ as defined in Algorithm \ref{alg:prototype},
$\Sigma_t = \frac{n}{\mathcal{N}}(I-K_t H_t)$ is a constant matrix given $\varphi_t$, $e_t\sim N(0, \epsilon_t \Sigma_t)$, and  the gradient term $\nabla_{\varphi} \log \pi(\varphi_{t-1}^a|\vz_{t})$ is given by 
$\nabla_{\varphi} \log \pi(\varphi_{t-1}^a|\vz_{t}) =\frac{\mathcal{N}}{n} \sum_{i=1}^n \nabla_{\varphi} \log \pi(z_t^{(i)}|\varphi_{t-1}^a)
+ \nabla_{\varphi} \log \pi(\varphi_{t-1}^a)$.
\end{lemma}
To establish the convergence of the preconditioned SGLD sampler (\ref{precondSGLD}), it suffices to establish the convergence of the conventional SGLD sampler in the context of reinforcement learning by noting the positive definiteness of the preconditioned matrix. Specifically, we have
\[
\small
\Sigma_t= \frac{n}{\mathcal{N}} (I-K_t H_t)=
\frac{n}{\mathcal{N}}[I-\epsilon_t H^T(\epsilon_t H_t H_t^T+R_t)^{-1} H_t],
\]
which implies $\Sigma_t$ has bounded positive eigenvalues for all $t \geq 1$. 


\subsubsection{Convergence of the LKTD algorithm under the on-policy setting} \label{onplicysect}
\vspace{-0.05in}

With a slight abuse of notations, we would prove the convergence of the following 
SGLD sampler in the RL context: 
\begin{equation} \label{SGLDsampler}
    \theta_k = \theta_{k-1} + \epsilon_k G(\theta_{k-1},\vz_k) + \sqrt{2 \beta^{-1}  \epsilon_k} \mathfrak{e}_k,
\end{equation}
where $\mathfrak{e}_k \sim N(0,I_d)$, $\beta$ is the inverse temperature, and 
$k$ indexes stages of the RL process. 
In pursuit of our objective, we introduce Assumption \ref{ass0} as detailed in the Appendix. 
This assumption aligns with the conditions outlined in \cite{Raginsky2017} 
to demonstrate the convergence of SGLD in simulating a posterior distribution with a 
fixed dataset. 
To tailor the sampler for RL, a context where the total sample size can be considered infinitely large, we introduce the pseudo-population size $\mathcal{N}$ to prevent the degeneration issue of the invariant distribution of $\theta_k$, thereby reformulating RL as a sampling problem rather than an optimization problem.
As a result, the proposed method can perform robustly with respect to the dynamics of the distribution $\pi(\bz|\theta_k)$.
However, under appropriate assumptions (see Remark \ref{rem:inference}), 
correct inference 
for the optimal policy can still be made based on the stationary distribution 
$\nu_{\mathcal{N}}(\theta) \propto \exp(-\beta \mathcal{G}(\theta))$, where 
$\mathcal{G}(\theta)=O(\mathcal{N})$ as stated in Theorem \ref{thm:LKTD}.
 Theorem \ref{thm:SGLD} and the followed Corollary \ref{thm:LKTD} establish the convergence of the LKTD algorithm under the general nonlinear setting for the value and Q-functions. Their proofs are given in Appendix. 

\begin{theorem} \label{thm:SGLD} 
Consider the SGLD sampler (\ref{SGLDsampler}) with a polynomailly-decay learning rate $\epsilon_k= \frac{\epsilon_0}{k^\varpi}$ for some $\varpi \in (0,1)$. 
Suppose the environment is stationary and Assumption \ref{ass0} holds. If $\mathbb{E} (G(\theta_{k-1},z_k))=g(\theta_{k-1})$ holds for any stage $k \in \{1,\dots, K\}$, $\beta \geq 1 \vee \frac{2}{m_U}$,  then there exist constants $(C_0, C_1, C_2, C_3)$ independent of the learning rates such that for all $K\in \mathbb{N}$, the 2-Wasserstein distance between
$\mu_K$ and $\nu_{\mathcal{N}}$ can be upper bounded by 
\begin{equation} \label{w2convergence}
\small
\begin{split}
    \mathcal{W}_2(\mu_K, \nu_{\mathcal{N}})
    &\leq  (12 + C_2  \epsilon_0(\frac{1}{1-\varpi}K^{1-\varpi}) )^\frac{1}{2} \cdot [( C_1\epsilon_0^2(\frac{2\varpi}{2\varpi-1}) + \delta C_0(\frac{ \epsilon_0}{1-\varpi}K^{1-\varpi}))^\frac{1}{2} \\
    & + ( C_1\epsilon_0^2(\frac{2\varpi}{2\varpi-1}) + \delta C_0(\frac{\epsilon_0}{1-\varpi}K^{1-\varpi}))^\frac{1}{4}] + C_3 \exp (-\frac{1}{\beta c_{LS}}(\frac{\epsilon_0}{1-\varpi}K^{1-\varpi})), 
\end{split}
\end{equation}
where $\mu_K(\theta)$ denotes the probability law of $\theta_K$,
$\nu_{\mathcal{N}}(\theta) \propto \exp(-\beta \mathcal{G}(\theta))$,
$\mathcal{G}(\theta)=O(\mathcal{N})$ is the anti-derivative of $g(\theta)$, i.e., 
$\nabla_{\theta} \mathcal{G}(\theta)=g(\theta)$, 
and $c_{LS}$ denotes a logarithmic Sobolev constant 
satisfied by the $\nu_{\mathcal{N}}$.  In addition, the constants $(C_0, C_1, C_2, C_3)$ are given by 
\[
\small
\begin{split}
C_0&=L_U^2(\kappa_0+ 2(1\vee\frac{e}{m_U}  )(b+2B^2+\frac{d}{\beta}))+B^2, \\
C_1 &=6L_U^2(C_0+\frac{d}{\beta}), \quad C_2 =\kappa_0 + 2b + 2d, \\
C_3 &= \sqrt{2c_{LS} (\log\|\nu_0\|_\infty + \frac{d}{2} \log\frac{3\pi}{m_U \beta} + \beta (\frac{M_U \kappa_0}{3} + B\sqrt{\kappa_0} + A + \frac{b}{2}\log 3) )}.  \\
\end{split}
\]
\end{theorem}

Regarding statistical inference with the samples $\{\theta_k: k=1,2,\ldots, K\}$, we have the remark: 

\begin{remark} \label{rem:inference}
Let $\phi(\theta))$ be a test function, which is bounded and differentiable. Suppose that the conditions of Theorem \ref{thm:SGLD} hold and $(\nu_{\mathcal{N}}(\theta), \phi(\theta))$ satisfies the Laplace regularity condition as given in Theorem 2.3 of \cite{SunSLiang2022sparseDNN}. Then, by Lapalce approximation, we have 
 \begin{equation} \label{Laplaceeq}
 \small
 \bar{\phi}_{\mathcal{N}}(\theta)=\frac{\int \phi(\theta) \exp(-\beta \mathcal{G}(\theta)) d \theta }{ \int \exp(-\beta \mathcal{G}(\theta)) d \theta} = \phi(\theta^*)+O(\frac{r_n^4}{\mathcal{N}}), \quad \mbox{as $K \to \infty$},
 \end{equation}
 where $\theta^*$ denotes the maximizer of $\nu_{\mathcal{N}}(\theta)$ and thus 
 the maximizer of $\nu_{\infty}(\theta)$,  and 
 $r_n$ denotes the connectivity of the sparse DNN learned for approximating 
 the value or $Q$-function. Therefore, when the number of total time steps $K$ becomes large and $\mathcal{N} \succ r_n^4$ holds, we can make inference for the policy using the Monte Carlo average
 $\hat{\phi}=[\sum_{k=1}^K \epsilon_k \phi(\theta_k)]/[\sum_{k=1}^K \epsilon_k]$,
 which, by letting $K \to \infty$, forms a  consistent estimator for $\phi(\theta^*)$, 
 the true value of $\phi(\cdot)$ at the optimal policy.  
\end{remark}
\begin{remark} \label{rem:N}
    The choice of the pseudo population size $\mathcal{N}$ reflects our trade-off between optimization and sampling. It acts as a tempering factor for the system. As $\mathcal{N} \to \infty$, we have $\mathcal{G}(\theta)\rightarrow \infty$ and, consequently, the stationary distribution $\nu_\mathcal{N}(\theta)$ degenerates to a delta function centered at $\theta^*$, as defined in Remark \ref{rem:inference}.
\end{remark}
We note that the proof for the convergence of $\hat{\phi}$ toward the population mean $\bar{\phi}_{\mathcal{N}}(\theta)$ requires that the learning rate sequence  satisfies 
Assumption \ref{ass:learning_rate}. This condition is readily met by the 
polynomailly-decay learning rate sequence outlined in Theorem \ref{thm:SGLD}.
Finally, we note that the conclusions of Theorem \ref{thm:SGLD} remains valid 
for Algorithm \ref{alg:LKTD}. Therefore, Remark \ref{rem:inference} and Remark \ref{rem:N} also hold for this algorithm.  
\begin{corollary} \label{thm:LKTD}
The conclusions of Theorem \ref{thm:SGLD} and Remark \ref{rem:inference} remains valid for Algorithm \ref{alg:LKTD}.
\end{corollary}


\subsubsection{Cooperation with Replay Buffer} \label{offpolicysect}

In Section \ref{onplicysect}, we demonstrated the $\mathcal{W}_2$-convergence of LKTD algorithm under an on-policy framework, where transition tuples $\vz_t$ are determined by the preceding parameter $\theta_{t-1}$. In contrast, off-policy algorithms obtain $\vz_t$ from replay buffers, a strategy that boosts data efficiency and is frequently adopted in Q-learning algorithms. This section is dedicated to establishing a convergence theory for the LKTD algorithm when it is integrated with a replay buffer. 
In practice, the replay buffer $\mathcal{B}$ stores transition $\{z_{t,j}\}_{j=1}^J$ drawn from the stationary distribution $\pi(z|\theta_{t-1})$ at each time step $t$. Given its finite capacity, the replay buffer retains only the transition tuples from the last $R$ time steps. That is, the replay buffer at time step $t$, denoted by $\mathcal{B}_t$, contains only the transition tuples generated from $\{\pi(z|\theta_\tau)\}_{\tau=t-R}^{t-1}$.  In the population scope, we define the replay buffer at time step $t$, denoted by $\bar{\pi}(z|\boldsymbol{\theta}_{t-1}^R)$, as a mixture of finite number of stationary distributions $\{\pi(z|\theta_{t-i})\}_{i=1}^R$. The replay buffer can be explicitly written as
\begin{equation}
 \small
    \bar{\pi}(z|\boldsymbol{\theta}_{t-1}^R) = \frac{1}{R} \sum_{i=1}^R
    \pi(z|\theta_{t-i}),
\end{equation}
 where $\boldsymbol{\theta}_{t-1}^R = \{\theta_{t-i}\}_{i=1}^R$ and $R\in \mathbb{N}$. Given the buffer's structure, we assume that the samples drawn from it are $R$-dependent. That is, $z_t$ and $z_{t'}$ are independent for all $z_t\in\mathcal{B}_t$, $z_{t'}\in\mathcal{B}_{t'}$ and $|t-t'|>R$. For some test function $\phi(\theta)$ of interest, we define the posterior average as: $\bar{\phi} = \int_{\Theta}\phi(\theta) \nu_{\mathcal{N}}(\theta)d\theta$, where $\nu_{\mathcal{N}}(\theta)$ is the target distribution as defined in Theorem \ref{thm:SGLD}. Let $\{\theta_t\}_{t=1}^T$ be the samples generated from LKTD algorithm, and the sample average $\hat{\phi}$ is as defined in Remark \ref{rem:inference}.
In Theorem \ref{thm:buffer}, we show that although the gradient is biased due to replay buffer, the bias and variance of $\hat{\phi}$ vanish asymptotically. In other words, we can employ replay buffer to improve data-efficiency without losing the asymptotic consistency.

\begin{theorem} \label{thm:buffer}
    Let $\{\theta_t\}_{t=1}^T$ be a sequence of updates generated from LKTD with replay buffer. At each time $t$, the transition tuple $z_t$ is sampled from the replay buffer $\bar{\pi}(z_t|\boldsymbol{\theta}_{t-1}^R)$. In addition to the assumptions in theorem \ref{thm:SGLD}, we further assume the following holds:
\begin{enumerate}[label=(\roman*)]
    \vspace{-0.05in}
    \item (Lipschitz) $\int_{\mathcal{Z}}|\pi(z|\theta) - \pi(z|\vartheta)|^2 dz \leq L\|\theta - \vartheta\|^2$;
    \vspace{-0.05in}
    \item (Integrability) $\int_{\mathcal{Z}} \|G(\theta, z)\|^2 dz \leq M$ and  $\int_{\mathcal{Z}} \|G(\theta, z)\|^2\pi(z|\theta) dz \leq M$, $\forall \theta\in \Theta$.
\end{enumerate}
Then for a bounded test function $\phi$, the bias of the LKTD can be bounded as:
\begin{equation}
\small
\begin{split}
    |\E\hat{\phi}-\bar{\phi}| =O(\frac{1}{S_T} + \frac{\sum_{t=1}^T \epsilon_t^2}{S_T}), \quad
    \E (\hat{\phi}-\bar{\phi})^2 =  O(\frac{1}{S_T} + \frac{\sum_{t=1}^T \epsilon_t^2 }{S_T^2} + \frac{(\sum_{t=1}^T \epsilon_t^2)^2}{S_T^2} )
\end{split}
\end{equation}
\end{theorem}
 
\section{Experiments}

This section compares LKTD with prominent RL algorithms such as DQN, BootDQN \citep{BootDQN}, QR-DQN \citep{bellemare2017distributional} and KOVA \citep{Shashua2020KalmanMB}. Using an simple indoor escape environment, we demonstrate the advantages of LKTD in three aspects: (1) accuracy of Q-value estimation, (2) uncertainty quantification of Q-values, and (3) optimal policy exploration. Furthermore, by employing more complex environments like OpenAI gym, we demonstrate that LKTD is capable of learning better and more stable policies for both training and testing. 

\begin{wrapfigure}{r}{0.275\textwidth}
         \vspace{-0.45in}
         \centering
         \includegraphics[width=0.25\textwidth]{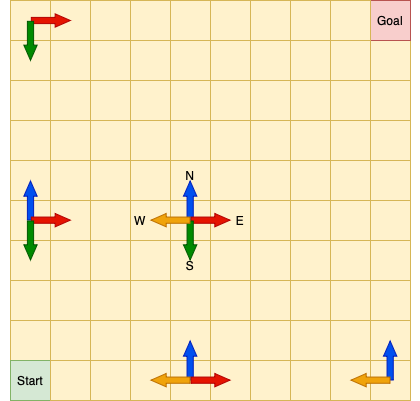}
         \caption{Indoor escape environment}
         \label{fig:indoor env}
         \vspace{-0.25in}
\end{wrapfigure}

\subsection{Indoor escape environment}\label{Indoor escape}

Figure \ref{fig:indoor env} depicts a simple indoor escape environment, for which the state space consists of 100 grids and the agent's objective is to navigate to the goal positioned at the top right corner. The agent starts its task from the bottom left grid at time $t=0$.
For every time step $t$, the agent identifies its current position, represented by the coordinate $s=(x,y)$. Based on a policy, the agent chooses an action $a\in\{\textup{N}, \textup{E}, \textup{S}, \textup{W} \}$. The action taken by the agent determines the adjacent grid to which it moves.
Following each action, the agent is awarded an immediate reward, $r_t$, drawn from the distribution $\mathcal{N}(-1, 0.01)$. It's worth noting that for most states, the Q-values for actions N and E are identical. This highlights the importance of exploring diverse optimal policies to achieve a consistent and resilient policy. Our experiment showcases the sampling framework are capable of learning a mixed optimal policy in a single run. 
We compare the proposed sampling framework LKTD against existing RL algorithms like DQN, BootDQN, QR-DQN and KOVA in the training of the deep Q-network. 
Refer to section \ref{appendix: indoor env} for the detailed experimental setup. 

For each algorithm, we collect the last 3000 parameter updates to form a  $\theta$-sample pool, denoted by $\boldsymbol{\theta}_s = \{\hat{\theta}_i\}$, 
which naturally induces a sample pool of Q-functions $\mathbf{Q}_s = \{ Q_\theta(\cdot, \cdot)| \theta\in \boldsymbol{\theta}_s\}$. We can obtain a point estimate of the Q-value at $(s,a)$ by calculating the sample average $\hat{Q}(s,a) = \frac{1}{n}\sum_{i=1}^n Q_{\hat{\theta}_i}(s,a)$. For uncertainty quantification, we can achieve one-step value tracking by constructing a 95\% prediction interval with the Q-value samples. 
Due to the simplicity of this environment, the Q-values of optimal policy, denoted by $Q^*(s,a)$, can be calculated by Monte Carlo simulations. 
For each algorithm and parameter setting, we conduct 100 runs and calculate two metrics: (1) the mean squared error (MSE) between $\hat{Q}(s,a)$ and $Q^*(s,a)$, denoted by MSE$(\hat{Q}_a)$ for each action $a$, where the average is taken over all states $s$; and (2) the coverage rate (CR) of the $95\%$ prediction intervals. 

\begin{figure}
    \vspace{-0.2in}
    \centering
    \includegraphics[width=0.95\textwidth]{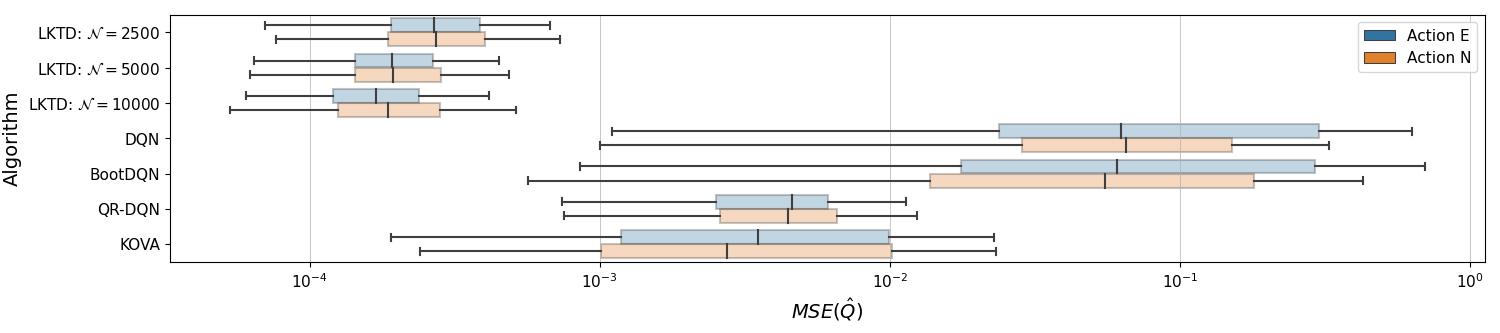}
    \vspace{-0.1in}
    \caption{Boxplots for MSE($\hat{Q}_a$) (for $a\in \{N,E\})$)} 
    \label{fig:mse_boxplot}
    \vspace{-0.1in}
\end{figure}

\begin{figure}
    \vspace{-0.05in}
    \centering
    \includegraphics[width=0.95\textwidth]{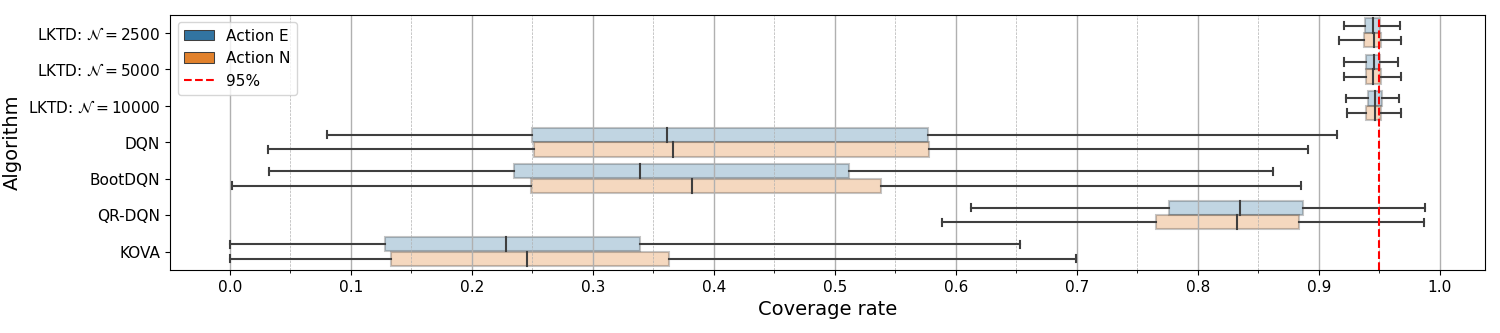}
    \vspace{-0.1in}
    \caption{Boxplots for coverage rates (for $a\in \{N,E\})$)}
    \label{fig:boxplot}
    \vspace{-0.2in}
\end{figure}

Figure \ref{fig:mse_boxplot} presents a boxplot illustrating the distribution of MSE($\hat{Q}_a$) (for $a\in \{N,E\}$) across 100 experiments for each algorithm. Here, we consider only the actions $a\in\{N, E\}$, since $\{S, W\}$ are sub-optimal actions at 
 all states and the corresponding Q-values cannot be well approximated due to the lack of enough transition tuples on them. Figure \ref{fig:mse_boxplot} indicates that LKTD yields notably higher Q-value estimation accuracy compared to all other algorithms. Moreover, the plot shows a clear trend that as the pseudo population size $\mathcal{N}$ grows, their accuracy correspondingly improves. For uncertainty quantification, Figure \ref{fig:boxplot} shows that the coverage rates from the LKTD algorithm is close to the nominal $95\%$ and independent of the choice of pseudo population size, whereas the DQN, BootDQN and KOVA algorithms fail to construct correct prediction intervals. Although QR-DQN approximates the distribution of the Q-function, it does not provide the correct interval estimation for Q-values.
 These observations on LKTD align well with the point we made in Remark \ref{rem:N}.

\begin{figure}[htbp]
 \vspace{-0.15in}
     \centering
        \begin{subfigure}[b]{0.30\textwidth}
         \centering
         \includegraphics[width=0.8\textwidth]{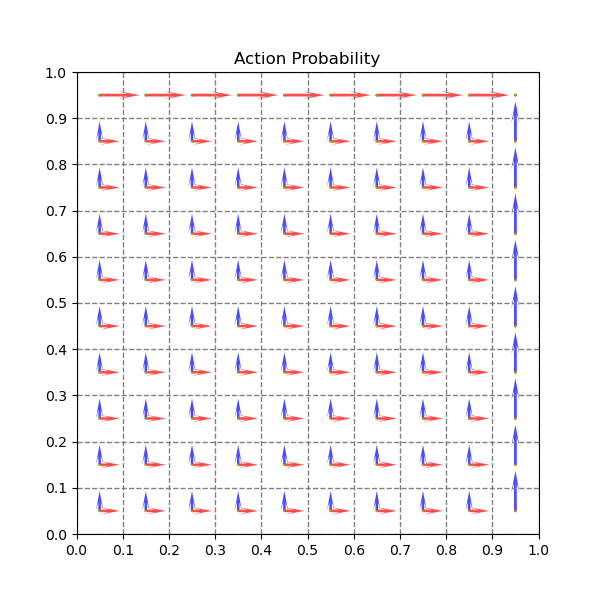}
         \caption{Optimal solution}
         \label{fig:mean optimal}
     \end{subfigure}
     \begin{subfigure}[b]{0.30\textwidth}
         \centering
         \includegraphics[width=0.8\textwidth]{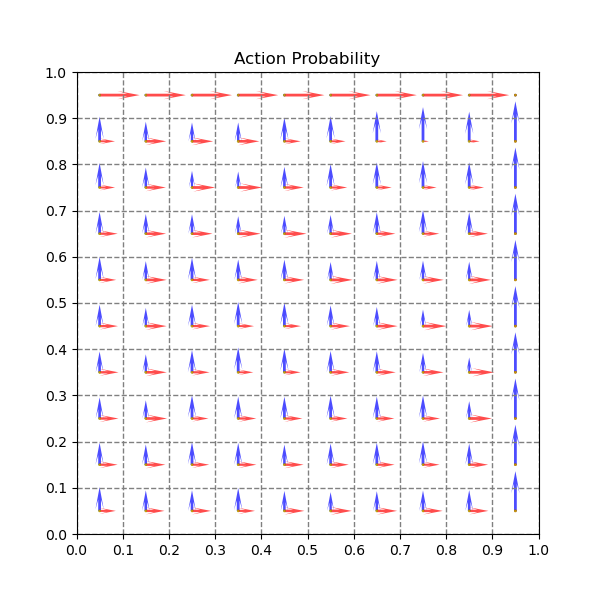}
         \caption{LKTD}
         \label{fig:action prob good}
     \end{subfigure}
     \begin{subfigure}[b]{0.30\textwidth}
         \centering
         \includegraphics[width=0.8\textwidth]{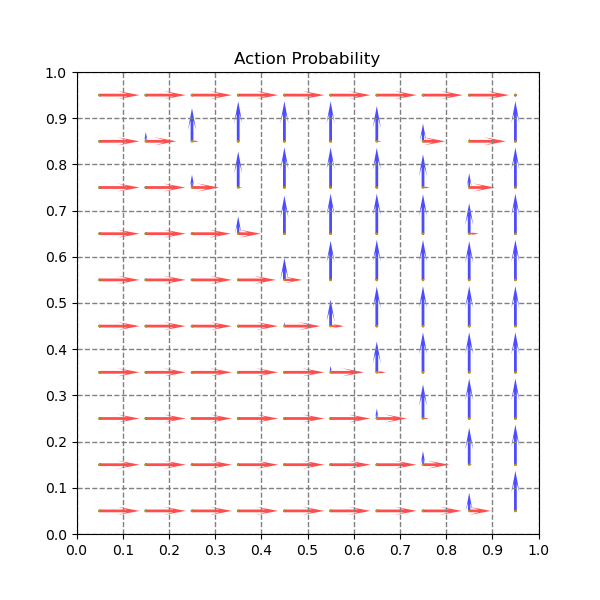}
         \caption{DQN}
         \label{fig:action prob bad}
     \end{subfigure}
       \vspace{-0.1in}
        \caption{ Mean policy probabilities for the indoor escape environment: (a) known optimal solution; (b) learned by LKTD; (c) learned by DQN, failing to explore different policies.}
        \label{fig:action prob}
 \vspace{-0.15in}
\end{figure}

Effective policy exploration is crucial for RL agents as it empowers them to adeptly learn various optimal policies and navigate challenges like the local-trap issue. In this specific environment, the Q-values for actions $N$ and $E$ are indistinguishable. Therefore, an algorithm excelling in policy exploration should, during training, give equal consideration to both optimal actions. To quantify policy exploration, we introduce the concept of \textbf{mean policy probability}. For a given policy pool $\varrho$ at state $s$, it's defined as:
$$
p_\varrho(a|s) = \frac{1}{|\varrho|}\sum_{\rho \in \varrho}\mathbf{1}_{a}(\rho(s)),
$$
where $\varrho$ represents the policy pool derived from  the $\theta$-sample pool obtained in a run of the algorithm. In simpler terms, the mean policy probability measures the frequency of an action chosen by the policy sample at a specific state.
Figure \ref{fig:action prob} shows that LKTD effectively explores the optimal policies across the majority of grids, whereas DQN sticks to one optimal policy, failing to explore others. This implies that DQN tends to be locally trapped in RL, compared to LKTD.

From a computational aspect, the LKTD algorithm stands out for its efficiency and scalability. As detailed in Table \ref{table:computation cost}, we have recorded the average computation time required by each algorithm to execute a single parameter update. The findings indicate that LKTD scale effectively in relation to network and batch size. Their time complexities align closely with that of DQN. Conversely, the KOVA algorithm, due to its reliance on the calculation of the Jacobian matrix and matrix inversion, proves to be computationally less efficient.

\subsection{Classical control problems}

\begin{figure}[htbp]
 \vspace{-0.1in}
    \centering
    \includegraphics[width=0.9\textwidth]{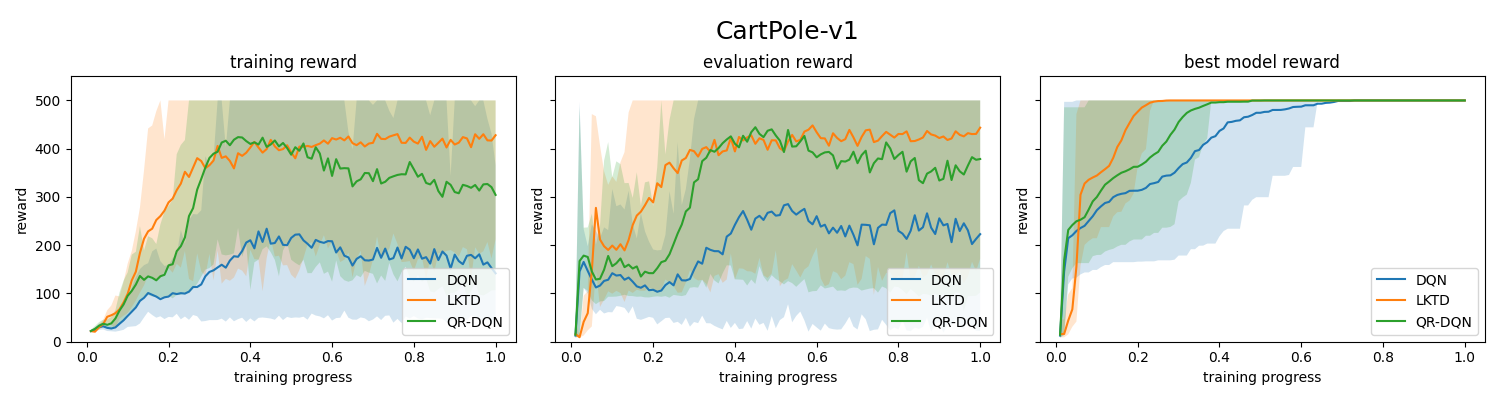}
    \vspace{-0.1in}
    \caption{CartPole-v1: The left plot shows the cumulative rewards obtained during the training process, the middle plot shows the testing performance without random exploration, and
    the right plot shows the performance of best model learned up to the point $t$.
    }
    \label{fig:cartpole}
    \vspace{-0.1in}
\end{figure}

This section evaluates LKTD's performance on four OpenAI gym challenges: Acrobot-v1, CartPole-v1, LunarLander-v2, and MountainCar-v0, comparing it against DQN and QR-DQN based on RL Baselines3 Zoo \citep{rl-zoo3} training framework. We conducted 100 replicates per experiment, with results in Figures \ref{fig:cartpole} and \ref{fig:gym}, excluding the top and bottom 5\% outliers.  Mean reward curves are shown with solid lines, and a 90\% confidence interval is indicated by the colored regions, showcasing LKTD's efficient exploration and robustness. For experimental details, see section \ref{sec: hyperparameter}.

Across all tested environments, LKTD consistently surpasses DQN and slightly better than QR-DQN. LKTD's ability to achieve markedly higher training reward indicates its capacity to learn more robust policies against sub-optimal actions generated by random exploration. This superiority is particularly pronounced in sensitive environments like CartPole-v1 (see Figure \ref{fig:cartpole}), where even a single misstep can lead to episode termination. Even when discounting the noise from random exploration, LKTD's evaluation rewards remain superior to DQN's. Additionally, when it comes to optimal policy exploration, LKTD can identify a more optimal model within an equivalent computational timeframe as DQN, highlighting its efficiency in policy exploration.

\section{Conclusion}


In this paper, we present a novel sampling framework designed to enhance SGD optimizers for addressing deep reinforcement learning challenges. By redefining the state-space model and introducing a pseudo population size, we enable SGMCMC algorithms, such as LKTD and SGLD, to converge to the accurate posterior distribution under mild conditions. Our approach outperforms existing value-based algorithms in benchmarks. Specifically, our LKTD algorithm demonstrates greater computational efficiency and circumvents potential matrix degeneration issues by eliminating the need for linearization, unlike the KOVA algorithm. Compared to DQN and its variants, our framework not only provides more precise point estimates of Q-values but also generates accurate prediction intervals for value tracking. In gym environments, LKTD surpasses DQN and QR-DQN in robustness and efficiency of optimal policy discovery. Overall, our framework signifies a significant advancement in deep RL optimization, offering improvements in both efficiency and precision.

\bibliography{reference}
\bibliographystyle{iclr2024_conference}

\newpage

\appendix
\section{Appendix}

\setcounter{table}{0}
\renewcommand{\thetable}{A\arabic{table}}
\setcounter{figure}{0}
\renewcommand{\thefigure}{A\arabic{figure}}
\setcounter{equation}{0}
\renewcommand{\theequation}{A\arabic{equation}}
\setcounter{lemma}{0}
\renewcommand{\thelemma}{A\arabic{lemma}}
\setcounter{theorem}{0}
\renewcommand{\thetheorem}{A\arabic{theorem}}
\setcounter{remark}{0}
\renewcommand{\theremark}{A\arabic{remark}}
\setcounter{algocf}{0}
\renewcommand{\thealgocf}{S\arabic{algocf}}
\setcounter{assumption}{0}
\renewcommand{\theassumption}{A\arabic{assumption}}

\subsection{Extended Kalman Temporal Difference algorithm}

 Let $\hat{\Sigma}_{t}$ denote the estimator for the covariance matrix of $\theta_{t}$. 
 Additionally, for equation (\ref{statespaceeq}), we 
 let $W_t \in \mathbb{R}^{p\times p}$ denote the covariance matrix of the Gaussian noise 
 $w_t$, and let $\Gamma_t \in \mathbb{R}^{n\times n}$ denote the covariance matrix of
 the Gaussian noise $\eta_t$. 
 The resulting KTD algorithm is Algorithm \ref{alg:KOVA} given in the Appendix. 
The major issues with Algorithm \ref{alg:KOVA} are (i) unknown approximation accuracy: 
it is unclear how well $N(\hat{\mu}_t, \hat{\Sigma}_t)$ approximates the true distribution
of $\theta_t$; (ii) computational complexity: it is $O(np^2)$ per iteration; and (iii) memory complexity: it requires  $O(p^2)$ additional space to store the covariance matrix 
$\hat{\Sigma}_t$.

\begin{algorithm}[htbp]
\SetAlgoLined
\textbf{Initialize}  $\hat{\mu}_0$\;  
 \For{t=1,2,\dots,T}{
   
   \textbf{(i)} Set predictions: $\hat{\mu}=\hat{\mu}_{t|t-1}=\hat{\mu}_{t-1}$ and 
   $\hat{\Sigma}_{t|t-1}=\hat{\Sigma}_{t-1}+W_{t}$\; 
   \textbf{(ii)} Generate $n$ transition tuples $\{\br_t,h(\bx_t,\hat{\mu}) \}$ from the system via the agent-environment interaction\;   
 \textbf{(iii)} Calculate $(p \times n)$-dim matrix $\nabla_{\theta} h(\bx_t,\hat{\mu})$ and the Kalman gain matrix 
\[
 K_t = \hat{\Sigma}_{t|t-1}  \nabla_{\theta} h(\bx_t,\hat{\mu}) \Gamma_{\tilde{\br}_t}^{-1},
\]
 where  $\Gamma_{\tilde{\br}_t} =  \nabla_{\theta} h(\bx_t,\hat{\mu})^T \hat{\Sigma}_{t|t-1} 
   \nabla_{\theta} h(\bx_t,\hat{\mu})+ \Gamma_t$\;  
   
 \textbf{(iv)} Update the mean and covariance matrix estimators: 
   \[
 \begin{split}
   \hat{\mu}_t &=\hat{\mu}_{t|t-1}+\tilde{\alpha} K_t(\br_t-h(\bx_t,\hat{\mu}_{t|t-1}), \\ 
   \hat{\Sigma}_{t} &= \hat{\Sigma}_{t|t-1}-\tilde{\alpha} K_t \Gamma_{\tilde{\br}_t} K_t^T, \\
  \end{split}
  \]
  where $\tilde{\alpha}$ is the learning rate. 
  }
\caption{Extended Kalman Temporal Difference Algorithm (KOVA Algorithm); \cite{Shashua2020KalmanMB}}
 \label{alg:KOVA}
\end{algorithm}

\subsection{The Prototype of the LKTD Algorithm}

By ignoring the detail of state augmentation, the model (\ref{eq:nonlinear state-space model}) can be simulated using Algorithm \ref{alg:prototype}, which is the single-chain version of Algorithm 2 of \cite{Zhang2021LEnKF}.  

\begin{algorithm}[htbp]
\SetAlgoLined
\textbf{Initialization:} Start with an initial parameter sample $\varphi_0^a\in \mathbb{R}^p$, drawn from the prior distribution $\pi(\varphi)$\;
\For{t=1,2,\dots,T}{   

     \textbf{Presetting:}
     Set $B_t = \epsilon_t I_p$, $R_t = 2 \sigma^2I_n$, and the Kalman gain matrix $K_{t} = B_t H_{t}^\top (H_{t} B_t H_{t}^\top + R_t)^{-1}$\;
     
    \textbf{Sampling:} With policy $\rho_{\varphi_{t-1}^a}$, generate a set of $n$ transition tuples from the stationary distribution $\mu_{\varphi_{t-1}^a}$ , denoted by $\vz_{t} = (\vr_t, \vx_t) =  \{z_{t,j}\}_{j=1}^n$\;

    \textbf{Forecast:} Draw $w_t\sim N_p(0,  \frac{n}{\mathcal{N}} B_t)$ and calculate
        \begin{equation}
            \varphi_t^{f} = \varphi_{t-1}^{a} + \frac{\epsilon_t}{2} \frac{n}{\mathcal{N}} \nabla \log \pi (\varphi_{t-1}^{a}) + w_t.
        \end{equation}

    \textbf{Analysis:} Draw $v_t\sim N_n(0, \frac{n}{\mathcal{N}} R_t)$ and calculate
        \begin{equation}
            \varphi_t^{a} = \varphi_{t}^{f} +K_{t}(\vr_{t} - H_{t} \varphi_t^{f} - v_t) := \varphi_{t}^{f} +K_t(\vr_{t} - \vr_t^{f }).
        \end{equation}
}
\caption{Prototype of the LKTD Algorithm}
\label{alg:prototype} 
\end{algorithm}

\subsection{Proofs for Theoretical Results}

\subsubsection{Assumptions}

 \begin{assumption} \label{ass0}~
\begin{itemize} 
\item[(C1)]  For any $\theta \in \Theta$, 
we are able to generate tuples $\vz$ from a unique stationary distribution $\pi(z|{\theta})$, the function $G:\Theta \times \mathcal{Z}$ is measurable, and $\|g(\theta)\|=\|\int_{\mathcal{Z}} G(\theta,z)\pi(z|\theta) dz \|< \infty$.

\item[(C2)] There exists a function $\mathcal{G}(\theta)$, which is an anti-derivative of $g(\theta)$ with respect to $\theta$, i.e., $\nabla_{\theta}\mathcal{G}(\theta)=g(\theta)$, such that $|\mathcal{G}(0)|\leq A$ for some constant $A>0$; in addition, there exists some constant $B>0$ such that 
$\|g(0)\| \leq B$. 

\item[(C3)] There exists some constant $L_U>0$ such that 
\[
 \|g(\theta)-g(\vartheta) \| \leq L_U \|\theta-\vartheta\|, \quad \forall  \theta,\vartheta \in \Theta.
\]
\item[(C4)]  The function $\mathcal{G}(\theta)$ is $(m_U,b)$-dissipative; that is, for some $m_U>0$ and $b \geq 0$, 
\[
\langle \theta, g(\theta) \rangle \geq m_U\|\theta\|^2 -b, \quad \forall \theta\in \Theta.
\]

\item[(C5)] There exist a constant $\delta$ and some constants $M_U$ and $B$ such that
\[
 \mathbb{E} \|G(\theta,z)-g(\theta)\|^2 \leq 2\delta(M_U^2 \|\theta\|^2+B^2), \quad \forall \theta\in \Theta.
 \]
  where the expectation is taken with respect to $z\sim \pi(z|\theta)$.
 \item[(C6)] The probability law $\mu_0$ of the initial hypothesis $\theta_0$ has a bounded and strictly positive density $p_0$ with respect to the Lebesgue measure on $\Theta$, and 
 \[
  \kappa_0:= \log \int_{\Theta} e^{\|\theta\|^2} p_0(\theta) d\theta < \infty.
 \]
\end{itemize} 
\end{assumption}

 In particular, the condition (C4) is quite standard for establishing 
the existence of an invariant distribution for $\theta$, see e.g. \cite{Raginsky2017} and 
\cite{XuPan2018}. It intuitively indicates that the dynamics stays inside a bounded domain in high probability; if $\theta_k$ is far away from the origin, then the dynamics are forced to get back around the origin. 

\begin{assumption} \label{ass:learning_rate}
 The learning rate sequence $\{\epsilon_t\}$ is decreasing, i.e., $0<\epsilon_{k+1}<\epsilon_k$, and satisfies that 
\[
i) \ \ \sum_{k=1}^\infty \epsilon_k = \infty; \quad ii) \lim_{K \rightarrow\infty} \frac{\sum_{k=1}^K \epsilon_k^{2}}{\sum_{k=1}^K \epsilon_k} = 0.  
\]
\end{assumption}

 \subsubsection{Proof of Theorem \ref{thm:SGLD}}
  
\begin{proof}
The proof of Theorem \ref{thm:SGLD} follows from Theorem 2 of \cite{zhang2020cyclical}. For convenience, we use $t$ and $k$ to index the continuous-time and discrete-time respectively. Firstly, we consider the following SDE
\begin{equation}
\label{eq: sde}
    d\theta_t = -g(\theta_t) dt + \sqrt{2}d\mathcal{W}_t,
\end{equation}
where  $g(\theta)=\int_\mathcal{Z} G(\theta,z)\pi(z|\theta)dz$. Let $\nu_t$ denote the distribution of $\theta_t$, and the stationary distribution of (\ref{eq: sde}) is denoted by $\nu_\infty$.
\begin{equation}
    \theta_{k+1} = \theta_{k} - \epsilon_{k+1} G(\theta_k, z_{k+1}) + \sqrt{2\epsilon_{k+1}\beta^{-1}} \mathfrak{e}_{k+1}.
\end{equation}
Further, let $\mu_k$ denote the distribution of $\theta_k$ and $S_k = \sum_{i=1}^k\epsilon_i$. Since 
\begin{equation}
    W_2(\mu_K,\nu_\infty) \leq W_2(\mu_K, \nu_{S_K}) + W_2(\nu_{S_K}, \nu_\infty),
\end{equation}
we need to bound these two terms respectively.\\
For the first term, $W_2(\mu_K, \nu_{S_K})$, our proof is based on the proof of Theorem 2 in \citet{zhang2020cyclical} with some modifications on learning rates. By definition, if $z_{k+1}$ is sampled from the stationary distribution $\pi(z|\theta_k)$, then $G(\theta_k, z_{k+1})$ is an unbiased estimator of $g(\theta_k)$, i.e., $\E[G(\theta_k, z_{k+1})|\mathcal{F}_k] = g(\theta_k)$, $\forall \theta_k\in \Theta\subset \mathbb{R}^d$. And we define $p(t)$ which will be used in the following proof:
\begin{equation}
    p(t) = \{k\in \mathbb{Z} | S_k \leq t < S_{k+1} \}
\end{equation}
Then we focus on the following continuous-time interpolation of $\theta_k $:
\begin{equation}
    \underline{\theta}(t) = \theta_0 - \int_0^t G(\underline{\theta}(S_{p(s)}), z_{p(s)+1}) ds + \sqrt{\frac{2}{\beta}} \int_0^t d\mathcal{W}_s^{(d)}
\end{equation}
where $G\equiv G_k$ for $t\in [S_k, S_{k+1})$. And for each $k$, $\underline{\theta}(S_k)$ and $\theta_k$ have the same probability law $\mu_k$. Since $\underline{\theta}(t)$ is not a Markov process, we define the following process which has the same one-time marginals as $\underline{\theta}(t)$
\begin{equation}
    V(t) = \theta_0 - \int_0^t H_s(V(s))ds + \sqrt{\frac{2}{\beta}} \int_0^t d\mathcal{W}_s^{(d)}
\end{equation}
with 
\begin{equation}
    H_t(x) := \E\left[ G( \underline{\theta}(S_{p(t)}) , z_{p(t)+1})| \underline{\theta}(t)=x \right]
\end{equation}
Let $\mathbf{P}_V^t := \mathcal{L}(V(s):0\leq s \leq t)$ and $\mathbf{P}_\theta^t := \mathcal{L}(\theta(s):0\leq s \leq t)$ and according to the proof of Lemma 3.6 in \citet{Raginsky2017}, we can derive a similar result for the relative entropy of $\mathbf{P}_V^t$ and $\mathbf{P}_\theta^t$:
\begin{equation}
\begin{split}
        D_{KL}(\mathbf{P}_V^t || \mathbf{P}_\theta^t) &= -\int d\mathbf{P}_V^t \log \frac{d\mathbf{P}_\theta^t}{d\mathbf{P}_V^t} \\
        &= \frac{\beta}{4} \int_0^t \E \| g(V(s)) - H_s(V(s)) \|^2 ds \\
        &= \frac{\beta}{4} \int_0^t \E \| g(\underline{\theta}(s)) - H_s(\underline{\theta}(s)) \|^2 ds \\
\end{split}
\end{equation}
The last line follows the fact that $\mathcal{L}(\underline{\theta}(s))=\mathcal{L}(V(s))$, $\forall s$. Then we will let $t = \sum_{k=1}^K \epsilon_k$ and we can use the martingale property of the integral to derive:
\begin{align}
        D_{KL}(\mathbf{P}_V^{\sum_{k=1}^K\epsilon_k} || \mathbf{P}_\theta^{\sum_{k=1}^K\epsilon_k}) 
        &= \frac{\beta}{4} \sum_{j=0}^{K-1} \int_{S_{j}}^{S_{j+1}} \E \|g(\underline{\theta}(s)) - H_s(\underline{\theta}(s))\|^2 ds \notag\\
        &= \frac{\beta}{2} \sum_{j=0}^{K-1} \int_{S_{j}}^{S_{j+1}} \E \|g(\underline{\theta}(s)) - g(\underline{\theta}(S_j))\|^2 ds \notag\\
        &\qquad + \frac{\beta}{2} \sum_{j=0}^{K-1} \int_{S_{j}}^{S_{j+1}} \E \|g(\underline{\theta}(S_j)) - H_s(\underline{\theta}(S_j))\|^2 ds \notag\\
        &= \frac{\beta L_U^2}{2} \sum_{j=0}^{K-1} \int_{S_{j}}^{S_{j+1}} \E \|\underline{\theta}(s) - \underline{\theta}(S_j)\|^2 ds \label{eq:term1}\\
        &\qquad + \frac{\beta}{2} \sum_{j=0}^{K-1} \int_{S_{j}}^{S_{j+1}} \E \|g(\underline{\theta}(S_j)) - H_s(\underline{\theta}(S_j)\|^2 ds \label{eq:term2}
\end{align}
For the ﬁrst part (\ref{eq:term1}), we consider some $s\in [ S_j,  S_{j+1} )$, for which the following holds:
\begin{equation}
    \begin{split}
        \underline{\theta}(s) - \underline{\theta}(S_j) &= -(s-S_j)G(\theta_k, z_{k+1}) + \sqrt{\frac{2}{\beta}} (\mathcal{W}_s^{(d)}-\mathcal{W}_{S_j}^{(d)}) \\
        &= -(s-S_j)g(\theta_k) + (s-S_j)(g(\theta_k) - G(\theta_k, z_{k+1})) + \sqrt{\frac{2}{\beta}} (\mathcal{W}_s^{(d)}-\mathcal{W}_{S_j}^{(d)})
    \end{split}
\end{equation}

Thus, we can use Lemma 3.1 and 3.2 in \citet{Raginsky2017} for the following result:
\begin{equation}
    \begin{split}
        \E \| \underline{\theta}(s) - \underline{\theta}(S_j) \|^2  &\leq 3\epsilon_{j+1}^2 \E\|g(\theta_j)\|^2 + 3\epsilon_{j+1}^2 \E\|g(\theta_j) - G(\theta_j, z_{j+1})\|^2  + 6\epsilon_{j+1} d \\
        &\leq 12 \epsilon_{j+1}^2 (L_U^2 \E\|\theta_j\|^2 + B^2) + \frac{6\epsilon_{j+1}d}{\beta} 
    \end{split}
\end{equation}
Hence we can bound the ﬁrst part, (choosing $\epsilon_0\leq 1$),
\begin{align}
    \frac{L_U^2}{2} \sum_{j=0}^{K-1} \int_{S_j}^{S_{j+1}} \E\| \underline{\theta}(s) - \underline{\theta}(S_{j}) \|^2 ds 
    &\leq \frac{L_U^2}{2} \sum_{j=0}^{K-1} [12\epsilon_{j+1}^3 (L_U^2\E\|\theta_j\|^2 + B^2) + \frac{6\epsilon^2_{j+1}d}{\beta}] \notag\\
    &\leq L_U^2 (\sum_{j=0}^{K-1} \epsilon_{j+1}^2)\max_{0\leq j\leq K-1} [6 (L_U^2 \E\|\theta_j\|^2 + B^2) + \frac{3d}{\beta} ] \\
    &\leq L_U^2 (\frac{\pi^2}{6}\epsilon_0^2) \max_{0\leq j\leq K-1} [6 (L_U^2 \E\|\theta_j\|^2 + B^2) + \frac{3d}{\beta} ] 
\end{align}
The second part (\ref{eq:term2}) can be bounded as follows:
\begin{align*}
    \frac{1}{2} \sum_{j=0}^{K-1} \int_{S_j}^{S_{j+1}} \E\| g(\underline{\theta}(S_j)) - H_s(\underline{\theta}(S_j))\|^2 ds
    &= \frac{1}{2} \sum_{j=0}^{K-1} \epsilon_{j+1} \E\| g(\theta_j) - G(\theta_j, z_{j+1}) \|^2\\
    &\leq \delta S_K \max_{0\leq j\leq K-1} (L_U^2 \E\|\theta_j\|^2 + B^2) \\
    &\leq \delta \epsilon_0(1+\log(K)) \max_{0\leq j\leq K-1} (L_U^2 \E\|\theta_j\|^2 + B^2) \\
\end{align*}
Due to the data-processing inequality for the relative entropy, we have
\begin{align*}
    & D_{KL}(\mu_K\| \nu_{S_K}) \leq D_{KL}(\mathbf{P}_V^t\|\mathbf{P}_\theta^t) \\
    &\leq \frac{L_U^2}{2} \sum_{j=0}^{K-1} \int_{S_j}^{S_{j+1}} \E\| \underline{\theta}(s) - \underline{\theta}(S_j) \|^2 ds + 
     \frac{1}{2} \sum_{j=0}^{K-1} \int_{S_j}^{S_{j+1}} \E\| g(\underline{\theta}(S_j)) - H_s(\underline{\theta}(S_j)) \|^2 ds \\
    &\leq L_U^2 (\sum_{j=0}^{K-1} \epsilon_{j+1}^2)\max_{0\leq j\leq K-1} [6 (L_U^2 \E\|\theta_j\|^2 + B^2) + \frac{3d}{\beta} ] + \delta S_K \max_{0\leq j\leq K-1} (L_U^2 \E\|\theta_j\|^2 + B^2) \\
    &\leq L_U^2 \epsilon_0^2(\frac{2\varpi}{2\varpi-1})\max_{0\leq j\leq K-1} [6 (L_U^2 \E\|\theta_j\|^2 + B^2) + \frac{3d}{\beta} ] \\
    & \quad + \delta \epsilon_0(\frac{1}{1-\varpi}K^{1-\varpi}) \max_{0\leq j\leq K-1} (L_U^2 \E\|\theta_j\|^2 + B^2) 
\end{align*}
According to the proof of Lemma 3.2 in \citet{Raginsky2017}, we can bound the term $\E\|\theta_k\|^2$
\begin{equation*}
    \E\|\theta_{k+1}\|^2 \leq (1-2\epsilon_{k+1} m_U + 4 \epsilon_{k+1}^2M_U^2) \E\|\theta_k\|^2 + 2 \epsilon_{k+1} b + 4\epsilon_{k+1}^2 B^2 + \frac{2\epsilon_{k+1} d}{\beta}
\end{equation*}
Similar to the statement of Lemma 3.2 in \citet{Raginsky2017}, we can fix $\epsilon_0 \in (0, 1\wedge \frac{m_U}{4M_U^2}\wedge \frac{1}{m_U})$. Then, we can know that
\begin{equation}
    \label{eq: theta bound}
    \E\|\theta_{k+1}\|^2 \leq (1-\epsilon_{k+1} m_U ) \E\|\theta_k\|^2 + 2 \epsilon_{k+1} (b + 2 B^2 + \frac{d}{\beta})
\end{equation}
where $\epsilon_K$ is the minimum of the decreasing learning rate sequence.
There are two cases to consider.
\begin{itemize}
    \item If $1-2\epsilon_K m_U + 4 \epsilon_K^2 M_U^2 \leq 0$, then from (\ref{eq: theta bound}) it follows that
    \begin{align*}
        \E\|\theta_{k+1}\|^2 &\leq  2\epsilon_0 (b+B^2+\frac{d}{\beta})\\
        &\leq \E\|\theta_0\|^2 + 2(b+B^2+\frac{d}{\beta})
    \end{align*}
    \item If $0\leq 1-2\epsilon_K m_U + 4 \epsilon_K^2 M_U^2 \leq 1$, then iterating (\ref{eq: theta bound}) gives
    \begin{align*}
        \E\|\theta_k\|^2 &\leq (1-\epsilon_{k} m_U ) \E\|\theta_{k-1}\|^2 + 2 \epsilon_{k} (b + 2 B^2 + \frac{d}{\beta}) \\
        &\leq e^{-\epsilon_{k} m_U }\E\|\theta_{k-1}\|^2 + 2 \epsilon_{k} (b + 2 B^2 + \frac{d}{\beta}) \\
        &\leq  e^{-m_U S_k} \E\|\theta_0\|^2 + 2(b + 2 B^2 + \frac{d}{\beta}) \sum_{i=1}^k \epsilon_i e^{-m_U (S_k-S_i)}\\
        &\leq  \E\|\theta_0\|^2 + 2(b + 2 B^2 + \frac{d}{\beta}) e^{-m_US_k} \sum_{i=1}^k \epsilon_i e^{m_U S_{i}} \\
        &\leq  \E\|\theta_0\|^2 + 2(b + 2 B^2 + \frac{d}{\beta}) e^{-m_US_k} \cdot e^{m_U\epsilon_0}\int_{0}^{S_k} e^{m_U x}dx \\
        &\leq  \E\|\theta_0\|^2 + 2(b + 2 B^2 + \frac{d}{\beta}) e^{-m_US_k} \cdot e^{m_U\epsilon_0} (\frac{1}{m_U} e^{m_US_k} - \frac{1}{m_U}) \\
        &\leq  \E\|\theta_0\|^2 + 2(b + 2 B^2 + \frac{d}{\beta}) \frac{e^{m_U\epsilon_0}}{m_U}  \\
        &\leq  \E\|\theta_0\|^2 + 2(b + 2 B^2 + \frac{d}{\beta}) \frac{e}{m_U}  \\
    \end{align*}
\end{itemize}
Now, we have 
\begin{align*}
    \max_{0\leq j\leq K-1} (L_U^2\E\|\theta_j\|^2 + B^2) \leq (L_U^2(\kappa_0+ 2(1\vee\frac{e}{m_U}  )(b+2B^2+\frac{d}{\beta}))+B^2):= C_0
\end{align*}
We denote the $6L_U^2(C_0+\frac{d}{\beta})$ as $C_1$ and we can derive
\begin{equation*}
    D_{KL}(\mu_K\| \nu_{S_K})\leq C_1 \epsilon_0^2(\frac{2\varpi}{2\varpi-1}) + \delta C_0  \epsilon_0(\frac{1}{1-\varpi}K^{1-\varpi})
\end{equation*}
Then according to Lemma 3.3 in \citet{Raginsky2017}, if we denote $\kappa_0 + 2b + 2d$ as $C_2$, we can derive the following result:
\begin{align*}
    W_2(\mu_K, \nu_{S_K}) &\leq (12 + C_2 S_K)^\frac{1}{2} \cdot [D_{KL}(\mu_K\| \nu_{S_K})^\frac{1}{2} + D_{KL}(\mu_K\|\nu_{S_K})^\frac{1}{4}] \\
    &\leq  (12 + C_2  \epsilon_0(\frac{1}{1-\varpi}K^{1-\varpi}) )^\frac{1}{2} \cdot [( C_1\epsilon_0^2(\frac{2\varpi}{2\varpi-1}) + \delta C_0 \epsilon_0(\frac{1}{1-\varpi}K^{1-\varpi}))^\frac{1}{2} \\
    & \qquad\qquad\qquad\qquad\qquad\qquad + ( C_1\epsilon_0^2(\frac{2\varpi}{2\varpi-1}) + \delta C_0\epsilon_0(\frac{1}{1-\varpi}K^{1-\varpi}))^\frac{1}{4}]
\end{align*}
Now we derive the bound for $W_2(\nu_{S_K}, \nu_\infty)$. By following the results in \citet{Raginsky2017} that there exist some positive constants $(C_3, c_{LS})$,
\begin{equation*}
    W_2(\nu_{S_K}, \nu_\infty) \leq C_3 \exp (-\frac{S_K}{c_{LS}})
\end{equation*}
\end{proof}

\subsubsection{Proof of Corollary \ref{thm:LKTD}}
\begin{proof}
By Theorem 1 of \cite{Ma2015ACR}, Algorithm \ref{alg:LKTD} works as a pre-conditioned SGLD algorithm with the pre-conditioner $\Sigma_t$, and it has the same stationary distribution as the SGLD algorithm (\ref{SGLDsampler}). 
By (\ref{w2convergence}), we have the 2-Wasserstein distance convergence  
for algorithm (\ref{SGLDsampler}) under the given assumptions. 
Therefore, for Algorithm \ref{alg:LKTD}, we also have  
$\mathcal{W}_2(\mu_T, \nu_{\mathcal{N}})\to 0$ as $T \to \infty$ 
by noting that $\Sigma_t$ is positive definite for any $t$.
\end{proof}

\subsubsection{Proof of Theorem \ref{thm:buffer}}

\begin{proof}
    By following the proof in \cite{chen2015convergence}, we define the functional $\psi$ that solves the Poisson Equation:
    \begin{equation}
        \mathcal{L}\psi (\theta_t) = \phi(\theta_t) - \bar{\phi}
    \end{equation}
    And $\psi$ satisfies the following smoothness condition
\begin{assumption}
\label{assumption:smoothness}
    $\psi$ and its up to 3rd-order derivatives, $\mathcal{D}^k\psi$, are bounded by a function $\mathcal{V}$, i.e., $\|\mathcal{D}^k\psi\|\leq C_k\mathcal{V}^{p_k}$ for $k=(0,1,2,3)$, $C_k$, $p_k>0$. Furthermore, the expectation of $\mathcal{V}$ on $\{\theta_t\}$ is bounded: $\sup_{s\in(0,1)} \mathcal{V}^p (s\theta +(1-s)Y)\leq C(\mathcal{V}^p(\theta) + \mathcal{V}^p(Y))$, $\forall\theta$, $Y$, $p\leq \max\{2p_k\}$ for some $C>0$.
\end{assumption}

\begin{assumption}
\label{assumption:eigenvalues}
    Let $\Sigma_t$ be the preconditioner, and assume that $\lambda_{t,\ell} \leq \inf_k \lambda_{\min}(\Sigma_t) \leq \sup_k \lambda_{\max} (\Sigma_t) \leq \lambda_{t,u}$ for some $\lambda_{t,\ell}$ and $\lambda_{t,u}$, where $\lambda_{\max}(\cdot)$ and $\lambda_{\min}(\cdot)$ denote the largest and smallest eigenvalues, respectively.
\end{assumption}

    First let us denote 
    \begin{equation}
        \tilde{\mathcal{L}}_t = \Sigma_t G(\theta_{t-1}, z_t)\cdot \nabla_\theta + \frac{1}{2} \Sigma_t \Sigma_t^\top : \nabla_\theta\nabla_\theta^\top
    \end{equation}
    the local generator of LKTD with replay buffer, where $\ba\cdot \bb:= \ba^\top \bb$ is the vector inner product, $\bm{A}:\bm{B} := \textup{tr}\{\bm{A}^\top \bm{B}\}$ is the matrix double dot product. Furthermore, let $\mathcal{L}$ be the true generator of the LKTD without replay buffer, that is, replacing the stochastic gradient in $\tilde{\mathcal{L}}_t$ with the true gradient. As a result, we have the relation:
    \begin{equation}
        \tilde{\mathcal{L}}_t = \mathcal{L} + \Delta V_t,
    \end{equation}
    where $\Delta V_t :=(G(\theta_{t-1}, z_t)-g(\theta_{t-1}))^\top \Sigma_t \nabla_\theta$, where $g(\theta_{t-1}) = \int_{\mathcal{Z}}G(\theta_{t-1}, z) \pi(z|\theta_{t-1}) dz$ is the true gradient, and $G(\theta_{t-1}, z_t)$ is the stochastic gradient calculated using transition tuples sampled from the replay memory. By following the proof of Theorem 1 in \citet{Li2016PreconditionedSG}, we can derive the estimation error as follows:
    \begin{equation}
        \hat{\phi} - \bar{\phi} = \frac{\E \psi(\theta_t)-\psi(\theta_0)}{S_T} + \frac{1}{S_T} \sum_{t=1}^{T-1}(\E \psi(\theta_{t-1}) - \psi(\theta_{t-1})) + \sum_{t=1}^T \frac{\epsilon_t}{S_T} \Delta V_t \psi (\theta_{t-1}) + O( \frac{\sum_{t=1}^T \epsilon_t^2}{S_T})
    \end{equation}
    By taking expectation on both side, we derived the bias as:
    \begin{equation}
    \label{eq:bias}
        |\E\hat{\phi} - \bar{\phi}| = O(\frac{1}{S_T} + \frac{\sum_{t=1}^T \epsilon_t \|\E\Delta V_t\|}{S_T} + \frac{\sum_{t=1}^T \epsilon_t^2}{S_T} )
    \end{equation}
    To prove the consistency of $\hat{\phi}$, we need to bound the term $\frac{1}{S_T}\sum_{t=1}^T \epsilon_t \|\E\Delta V_t\|$. By Assumption \ref{assumption:smoothness} and \ref{assumption:eigenvalues}, it is sufficient to prove that $\sum_{t=1}^T \epsilon_t \|\E\zeta_t \|$ is bounded, where $\zeta_t = G(\theta_{t-1}, z_t)-g(\theta_{t-1})$ is the gradient bias at time $t$. Let $ \bar{g}(\boldsymbol{\theta}_{t-1}^R) := \int_{\mathcal{Z}}G(\theta_{t-1}, z)\bar{\pi}(z|\boldsymbol{\theta}_{t-1}^R) dz$ be the expectation of the biased gradient given the replay buffer $\bar{\pi}(z|\boldsymbol{\theta}_{t-1}^R)$. By applying assumption (i) and (ii), we can bound the conditional expectation of the gradient bias
    

    \begin{equation}
        \begin{split}
            \|\E[\zeta_t|\mathcal{F}_{t-1}] \|^2 &= \|\bar{g}(\boldsymbol{\theta}_{t-1}^R) - g(\theta_{t-1})\|^2 \\
             & \leq \| \int_\mathcal{Z} G(\theta_{t-1}, z) \cdot (\bar{\pi}(z|\boldsymbol{\theta}_{t-1}^R) - \pi(z|\theta_{t-1}) ) dz  \|^2 \\
             & \leq (\int_\mathcal{Z} \|G(\theta_{t-1}, z)\| \cdot |\bar{\pi}(z|\boldsymbol{\theta}_{t-1}^R) - \pi(z|\theta_{t-1}) | dz )^2\\
             & \leq \int_{\mathcal{Z}}\|G(\theta_{t-1}, z)\|^2 dz \cdot \int_{\mathcal{Z}} |\bar{\pi}(z|\boldsymbol{\theta}_{t-1}^R) - \pi(z|\theta_{t-1}) |^2 dz \\
             & \leq M  \int_{\mathcal{Z}} \frac{1}{R} \sum_{i=1}^R|\pi(z|\theta_{t-i}) - \pi(z|\theta_{t-1}) |^2 dz\\
             & \leq \frac{ML}{R} \sum_{i=1}^R \|\theta_{t-i} - \theta_t\|^2
        \end{split}
    \end{equation}

    By Jensen's inequality,
    \begin{equation}
        \label{eq:expected bias bound}
        \begin{split}
             \|\E [ \E[\zeta_t|\mathcal{F}_{t-1}]] \|^2 &\leq  \E\|\E[\zeta_t|\mathcal{F}_{t-1}] \|^2 \\
             & \leq \E[ \frac{ML}{R} \sum_{i=1}^R \|\theta_{t-i} - \theta_t\|^2 ] \\
             & \leq C \frac{1}{R}\sum_{i=1}^R\E\|\theta_{t-i} - \theta_{t-1}\|^2 \\
             & \leq C \frac{1}{R}\sum_{i=1}^R  \sum_{j=1}^{i-1} (i-1)\E\|\theta_{t-j} - \theta_{t-j-1}\|^2 
        \end{split}
    \end{equation}
    
    Now, we want to bound the expected square difference between parameter updates
    \begin{equation}
    \label{eq:parameter bound}
        \begin{split}
            \E\|\theta_{t} - \theta_{t-1}\|^2 &= \E\|\epsilon_t \Sigma_t G(\theta_{t-1}, z_t) + \sqrt{2\epsilon_t}e_t \|^2 \\
            & \leq 2\epsilon_t^2 \|\Sigma_t\|^2 \E\| G(\theta_{t-1}, z_t)\|^2 + 4\epsilon_t  \|\Sigma_t\| \\
            & \leq 2\epsilon_t^2 \lambda_{\max}^2 \E\| G(\theta_{t-1}, z_t)\|^2 + 4\epsilon_t  \lambda_{\max}\\
            & \leq O(\epsilon_t^2)
        \end{split}
    \end{equation}
    
    Combine the results in (\ref{eq:expected bias bound}) and (\ref{eq:parameter bound}), we have
    \begin{equation}
    \label{eq:gradient bias order}
        \begin{split}
            \|\E[\zeta_t]\|^2 &\leq  C\frac{1}{R}\sum_{i=1}^R  \sum_{j=1}^{i-1} (i-1)\E\|\theta_{t-j} - \theta_{t-j-1}\|^2  \\
            &\leq C\frac{1}{R}\sum_{i=1}^R  \sum_{j=1}^{i-1} (i-1) O(\epsilon_{t-j}^2) \\
            &\leq C\frac{R-1}{2} O(\epsilon_{t-R}^2)\\
            &\leq O(\epsilon_{t-R}^2) =  O(\epsilon_{t}^2)
        \end{split}
    \end{equation}
    By (\ref{eq:gradient bias order}), the operator norm $\|\E\Delta V_t\|$ is of order $O(\epsilon_t)$. Combining this result with (\ref{eq:bias}) we can derive the bound for the bias
    \begin{equation}
    \begin{split}
        |\E\hat{\phi} - \bar{\phi}| &= O(\frac{1}{S_T} + \frac{\sum_{t=1}^T \epsilon_t \|\E\Delta V_t\|}{S_T} + \frac{\sum_{t=1}^T \epsilon_t^2}{S_T} ) \\
        &= O(\frac{1}{S_T} + \frac{\sum_{t=1}^T \epsilon_t^2}{S_T} + \frac{\sum_{t=1}^T \epsilon_t^2}{S_T} ) \\
        &=  O(\frac{1}{S_T} + \frac{\sum_{t=1}^T \epsilon_t^2}{S_T}) 
    \end{split}
    \end{equation}
Now, consider the $L^2$ convergence of $\hat{\phi}$. Since $\E\Delta V_t$ is nonzero under the setting of replay buffer, we follow the proof of Theorem 3 in \citet{chen2015convergence} with some modification. The MSE of $\hat{\phi}$ can be written as 
\begin{equation}
\small
    \begin{split}
        \E (\hat{\phi} - \bar{\phi})^2 & \leq C\E \Big\{  \frac{ (\E \psi(\theta_t)-\psi(\theta_0))^2}{S_T^2} + \frac{\sum_{t=1}^{T-1}(\E \psi(\theta_{t-1}) - \psi(\theta_{t-1}))^2}{S_T^2} + \big(\sum_{t=1}^T \frac{\epsilon_t}{S_T} \Delta V_t \psi (\theta_{t-1})\big)^2 \\
        & \quad + C( \frac{\sum_{t=1}^T \epsilon_t^2}{S_T})^2  \Big\} \\
    \end{split}
\end{equation}
Let $\Delta V_t = \E\Delta V_t + \delta_t^\top \Sigma_t \nabla_\theta$, where $\delta_t = (G(\theta_{t-1}, z_t)-\bar{g}(\boldsymbol{\theta}_{t-1}^R)) $ has mean 0. Since $z_t$'s are $R$-dependent due to the structure of replay buffer, $Cov(\delta_t, \delta_{t'})=0$ for all $|t-t'|>R$.  By assumption (ii), $\E\|\delta_t\|^2$ is bounded. Hence, we can derive the following bound:

\begin{equation}
    \begin{split}
        \E\|\sum_{t=1}^T \frac{\epsilon_t}{S_T} \Delta V_t \|^2 
        & \leq 2\|\sum_{t=1}^T \frac{\epsilon_t}{S_T} \E\Delta V_t \|^2 + 2\E\| \sum_{t=1}^T \frac{\epsilon_t}{S_T} \delta_t \Sigma_t \nabla_\theta\|^2 \\
        & \leq 2 (\sum_{t=1}^T \frac{\epsilon_t^2}{S_T^2}) (\sum_{t=1}^T \| \E\Delta V_t \|^2) + 2C\sum_{|t-t'|< R} \frac{\epsilon_t\epsilon_{t'}}{S_T^2} Cov(\delta_t, \delta_{t'})\\
        & = O(\frac{(\sum_{t=1}^T \epsilon_t^2)^2}{S_T^2} + \frac{R\sum_{t=1}^T \epsilon_t^2 \E\|\delta_t\|^2}{S_T^2}) \\
        & = O(\frac{(\sum_{t=1}^T \epsilon_t^2)^2}{S_T^2} + \frac{\sum_{t=1}^T \epsilon_t^2 }{S_T^2})
    \end{split}
\end{equation}
Finally, we can derive the MSE as 
\begin{equation}
    \E(\hat{\phi} - \bar{\phi})^2 = O(\frac{(\sum_{t=1}^T \epsilon_t^2)^2}{S_T^2} + \frac{\sum_{t=1}^T \epsilon_t^2 }{S_T^2} +\frac{1}{S_T})
\end{equation}
\end{proof}

\subsection{More Numerical Results}
\subsubsection{Implementation of the SGLD and SGHMC algorithm}
The notion of pseudo population introduced in the proposed LKTD algorithm can also be applied to SGLD and SGHMC algorithm. As implied by Lemma \ref{lem:1}, we can directly implement equation (\ref{precondSGLD}) with $\Sigma_t$ being restricted to an identity matrix, which leaves the same stationary distribution.  The pseudocode of SGLD and SGHMC are given in 
Algorithm \ref{alg:SGLD} and Algorithm \ref{alg:SGHMC}, where $\mK$ is set to match the computation of LKTD.

\begin{algorithm}[htbp]
\SetAlgoLined
\textbf{Initialization:} Draw $\theta_0^a\in \mathbb{R}^p$ drawn from the prior distribution $\pi(\theta)$.

\For{t=1,2,\dots, T}{
    \textbf{Sampling:} With policy $\rho_{\theta_{t-1}^a}$, generate a set of $n$ transition tuples, 
    denoted by  $\vz_{t} = (\vr_t, \vx_t) :=  \{r_t^{(j)}, x_{t}^{(j)}\}_{j=1}^n$, where 
  $x_t^{(j)} = (s_t^{(j)}, a_t^{(j)}, s_{t+1}^{(j)}, a_{t+1}^{(j)})^T$  
   and $x_t^{(j)} = (s_t^{(j)}, a_t^{(j)}, s_{t+1}^{(j)})^T$ correspond to the 
   choices of the $Q$-function and  $V$-function in (\ref{h-function}), respectively.

 \For{k=1,2,\ldots,$\mathcal{K}$}{
      
       \textbf{Presetting:} Set $B_{t,k} = \epsilon_{t,k} I_{\tilde{p}}$.
       
    \textbf{Draw} $\tilde{w}_{t,k}\sim N_p(0, \frac{n}{\mathcal{N}} B_{t,k})$ and calculate
        \begin{equation}
            \theta_{t,k} = \theta_{t,k-1} + \frac{\epsilon_{t,k}}{2} \frac{n}{\mathcal{N}} \nabla_{\theta} \log \pi (\theta_{t,k-1}|\bz_t) + \tilde{w}_{t,k},
        \end{equation}
    where $\theta_{t,0} =\theta_{t-1,\mathcal{K}}$ if $k=1$, and the gradient term is given by 
    \begin{equation}
        \nabla_{\theta} \log \pi (\theta_{t,k-1}|\bz_t) = 
        \nabla_\theta \log \pi(\theta_{t,k-1}) + \frac{1}{\sigma^2} 
        \frac{\mathcal{N}}{n} \nabla_\theta h(\vx_t; \theta_{t,k-1})(\vr_t- h(\vx_t; \theta_{t,k-1})) 
    \end{equation}
} 
}
\caption{SGLD for RL sampling framework}
\label{alg:SGLD}
\end{algorithm} 

\begin{algorithm}[htbp]
\SetAlgoLined
\textbf{Initialization:} Draw $\theta_0^a\in \mathbb{R}^p$ drawn from the prior distribution $\pi(\theta)$, momentum coefficient $\alpha$.

\For{t=1,2,\dots, T}{
    \textbf{Sampling:} With policy $\rho_{\theta_{t-1}^a}$, generate a set of $n$ transition tuples, 
    denoted by  $\vz_{t} = (\vr_t, \vx_t) :=  \{r_t^{(j)}, x_{t}^{(j)}\}_{j=1}^n$, where 
  $x_t^{(j)} = (s_t^{(j)}, a_t^{(j)}, s_{t+1}^{(j)}, a_{t+1}^{(j)})^T$  
   and $x_t^{(j)} = (s_t^{(j)}, a_t^{(j)}, s_{t+1}^{(j)})^T$ correspond to the 
   choices of the $Q$-function and  $V$-function in (\ref{h-function}), respectively.  \\
   Set $v_{t,0}=0$\\
 \For{k=1,2,\ldots,$\mathcal{K}$}{
      
       \textbf{Presetting:} Set $B_{t,k} = \epsilon_{t,k} I_{\tilde{p}}$.
       
    \textbf{Draw} $\tilde{w}_{t,k}\sim N_p(0, \alpha\frac{n}{\mathcal{N}} B_{t,k})$ and calculate
        \begin{equation}
        \begin{split}
            v_{t,k} &= (1-\alpha) v_{t,k-1} + \frac{\epsilon_{t,k}}{2} \frac{n}{\mathcal{N}} \nabla_{\theta} \log \pi (\theta_{t,k-1}|\bz_t) + \tilde{w}_{t,k} \\
            \theta_{t,k} &= \theta_{t,k-1} +  v_{t,k}
        \end{split}
        \end{equation}
    where $\theta_{t,0} =\theta_{t-1,\mathcal{K}}$ if $k=1$, and the gradient term is given by 
    \begin{equation}\label{eq:nonlinear gradient2}
        \nabla_{\theta} \log \pi (\theta_{t,k-1}|\bz_t) = 
        \nabla_\theta \log \pi(\theta_{t,k-1}) + \frac{1}{\sigma^2} 
        \frac{\mathcal{N}}{n} \nabla_\theta h(\vx_t; \theta_{t,k-1})(\vr_t- h(\vx_t; \theta_{t,k-1})) 
    \end{equation}
} 
}
\caption{SGHMC for RL sampling framework}
\label{alg:SGHMC}
\end{algorithm}

\subsubsection{Indoor escaping environment} \label{appendix: indoor env}

This section serves as a complement to Section \ref{Indoor escape} in the main text, offering more comprehensive experiment settings and numerical results to compare SGMCMC sampling algorithms with non-sampling algorithms. The SGMCMC sampling algorithms considered comprise LKTD, SGLD, and SGHMC, while the non-sampling algorithms encompass DQN, BootDQN, QR-DQN, and KOVA.

In this experiment, the Q-function is approximated by a deep neural network with two hidden layers of sizes (32, 32). The agent updates the network parameters every 10 interactions, for a total of $10^6$ action steps. The replay buffer size is set to $10^4$. For action selection, we use $\epsilon$-greedy exploration with a final exploring rate of $\epsilon=0.01$. The batch size is 100. 
To achieve sparse deep neural network, we follow the suggestion in \cite{SunSLiang2022sparseDNN}, let the deep neural network parameters be subject to a mixture Gaussian prior: 
\begin{equation}
    \theta \sim  (1-\lambda) \mathcal{N}(0, \sigma_0^2) + \lambda \mathcal{N}(0, \sigma_1^2)
\end{equation}
where $\lambda\in (0,1)$  is the mixture proportion and $\sigma_0^2$ is usually set to a small number compare to $\sigma_1^2$. We set $\sigma_1=0.5$, $\sigma_0 = 0.05$ and $\lambda = 0.5$ in all SGMCMC simulations. In equation (\ref{statespaceeq2}), the reward $r_t$ is assumed to be a Gaussian distribution with variance $\sigma^2$. For indoor escape environment, the reward is given by 
$\mathcal{N}(-1, 0.01)$; that is, we set $\sigma^2 = 0.01$.

For BootDQN, the number of "heads" is configured to $10$, with the Bernoulli probability set at $0.5$. For QR-DQN, the return distribution is approximated by $10$ quantiles.

For this problem, the optimal policy is not unique, any policy that choose either action N or action E at any inner state is an optimal policy. Despite that there are multiple optimal policies, they all share the same Q-function, denoted by $Q^*(\cdot,\cdot)$. Since we adopt $\epsilon$-greedy exploration, we re-denoted $Q^*(\cdot,\cdot)$ by 
$Q_\eps^*(\cdot,\cdot)$ to indicates its dependence on  the $\epsilon$-greedy exploration strategy. 
For all state-action pairs $(s,a)$, the Q-value $Q^*_\epsilon(s,a)$ can be estimated by Monte Carlo simulations. Note that $Q^*_\epsilon(\cdot, \cdot)$ is the target function that the deep neural network is to approximate.  

For each algorithm, we collect from the last 3000 parameter updates to form a  $\theta$-sample pool, denoted by $\boldsymbol{\theta}_s = \{\hat{\theta}_i\}$, 
which naturally induces a sample pool of Q-functions $\mathbf{Q}_s = \{ Q_\theta(\cdot, \cdot)| \theta\in \boldsymbol{\theta}_s\}$. We can obtain a point estimate of the Q-value at $(s,a)$ by calculating the sample average $\hat{Q}(s,a) = \frac{1}{n}\sum_{i=1}^n Q_{\hat{\theta}_i}(s,a)$. For uncertainty quantification, we can achieve one-step value tracking by constructing a 95\% prediction interval with the Q-value sample pool.

\begin{figure}
    \centering
    \includegraphics[width=0.95\textwidth]{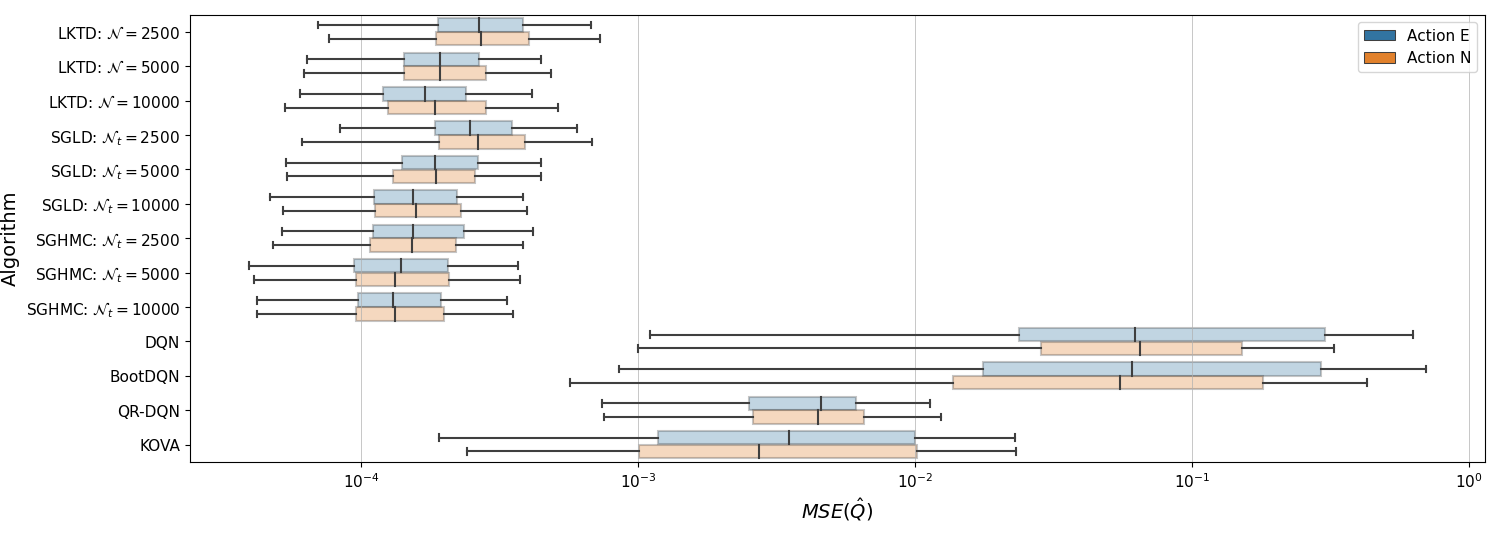}
    \vspace{-0.1in}
    \caption{Boxplots for MSE($\hat{Q}_a$) (for $a\in \{N,E\})$)} 
    \label{fig:mse_boxplot_full}
    \vspace{-0.1in}
\end{figure}

\begin{figure}
    \centering
    \includegraphics[width=0.95\textwidth]{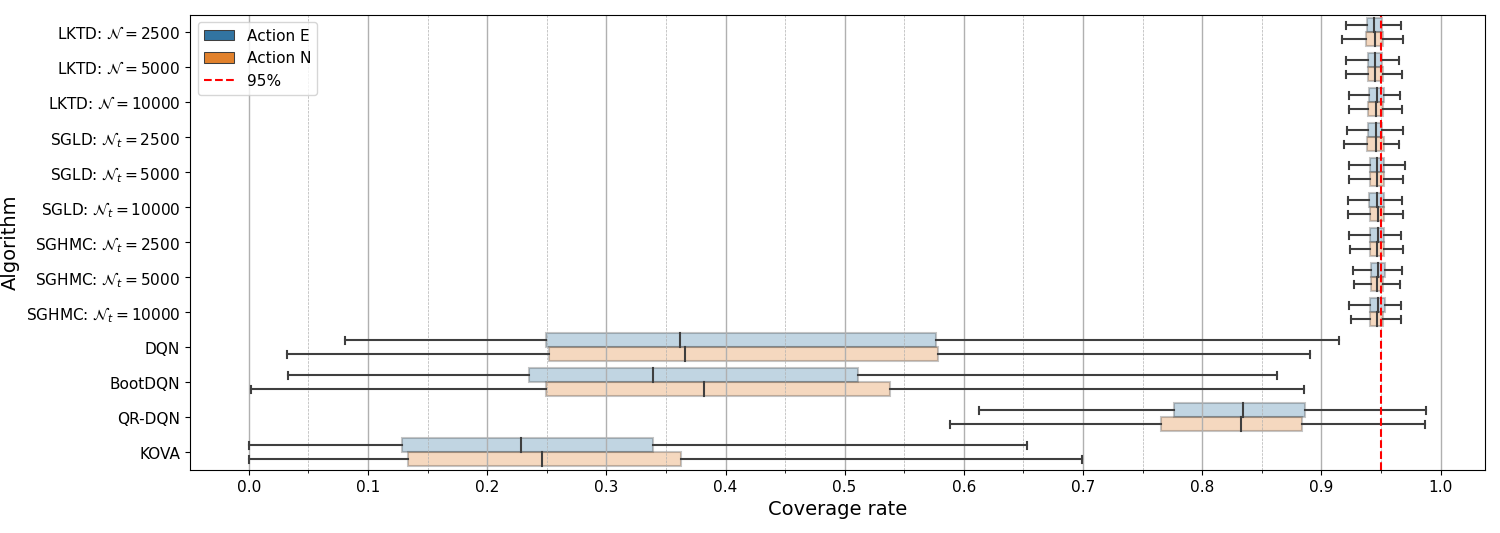}
    \vspace{-0.1in}
    \caption{Boxplots for coverage rates} 
    \label{fig:cr_boxplot_full}
    \vspace{-0.1in}
\end{figure}

For each algorithm and parameter setting, we conduct 100 runs and calculate two metrics for each action at each run: (i) the mean squared error (MSE) between $\hat{Q}(s,a)$ and $Q_\epsilon^*(s,a)$, denoted by MSE$(\hat{Q}_a)$, where the average is taken over all grids, that is, 
\[
\textup{MSE}(\hat{Q}_a)=\frac{1}{|\mathcal{S}|}\sum_{s\in\mathcal{S}} |\hat{Q}(s,a) - Q^*_\epsilon(s,a)|^2,
\]
 and (ii) the coverage rate (CR) of the $95\%$ prediction interval of $Q^*_\epsilon(s,a)$, that is, the probability of $Q^*_\epsilon(s,a)$ falling inside the prediction interval. 
 Figure \ref{fig:mse_boxplot_full} and Figure \ref{fig:cr_boxplot_full} show the boxplots of 
 $\textup{MSE}(\hat{Q}_a)$ (with $a\in \{N,E\}$) and coverage rates, respectively. In Figure \ref{fig:mse_boxplot_full}, the SGMCMC algorithms exhibit significantly smaller MSEs compared to other algorithms. It is worth noting that as the pseudo population increases, the MSE decreases, which supports the theoretical result in Remark \ref{rem:inference}. As shown in Figure \ref{fig:cr_boxplot_full}, the coverage rates of all SGMCMC algorithms achieve the nominal $95\%$ and independent of the choice of pseudo population size, whereas the DQN, BootDQN and KOVA algorithms fail to construct correct prediction intervals. Although the QR-DQN algorithm achieves a slightly higher coverage rate than DQN, the results prove that it does not converge to the correct return distribution.

In Table \ref{table:MSE Q mean}, we have recorded the trimmed mean (standard deviation) of MSE($\hat{Q}_a$) (for $a\in \{N,E\}$) 
over 100 runs, where trimmed means are calculated by excluding the outliers. The outliers are the values that falls outside the interval (Q1-1.5IQR, Q3+1.5IQR), where Q1 and Q3 are, resepctively, the 1st and 3rd quartiles of the samples, and IQR = Q3 - Q1. Both tables indicate that SGMCMC algorithms are more accurate than non-sampling algorithms in Q-function approximation. Regarding uncertainty quantification, Table \ref{table:CR mean} presents the trimmed mean of the coverage rates and lengths of prediction intervals. It is worth noting that the prediction interval shrinks as the pseudo population size increases, which aligns with our theory as mentioned in Remark \ref{rem:inference}.



\begin{figure}
    \centering
    \includegraphics[width=0.95\textwidth]{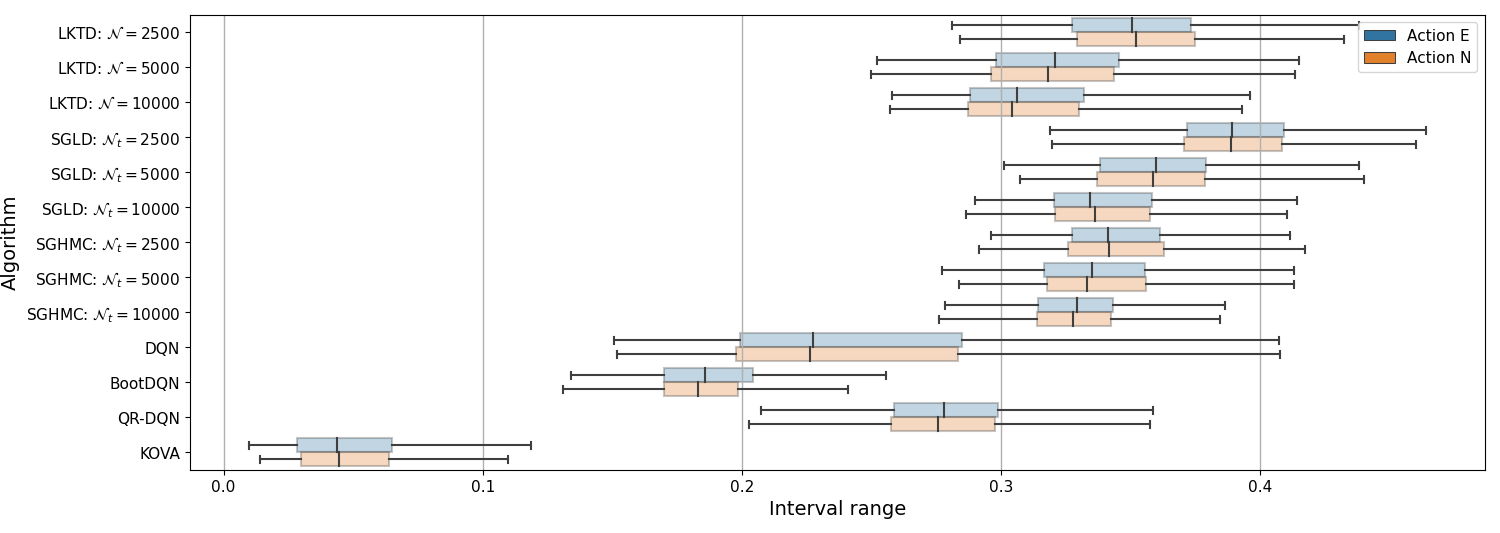}
    \vspace{-0.1in}
    \caption{Boxplots for the range of prediction intervals} 
    \label{fig:range_boxplot_full}
    \vspace{-0.1in}
\end{figure}

\begin{table}[htbp]
  \caption{Trimmed mean of MSE($\hat{Q}_a$) ($a \in \{N,E\}$) over 100 runs}
  \label{table:MSE Q mean}
  \centering
  \adjustbox{max width=\textwidth}{
  \begin{tabular}{lcccccccccc}
    \toprule
    Algorithm     &  $\epsilon_t$  & $\mathcal{N}$   & East  & North  \\
    \midrule
    LKTD  & 1e-5   & 2500   &  0.00031 (0.00018) & 0.00035 (0.00026) \\
    LKTD  & 1e-5   & 5000   &  0.00023 (0.00015) & 0.00023 (0.00015) \\
    {\bf LKTD}  & \bf {1e-5}   & \bf {10000}  & {\bf 0.00020 (0.00012)} & {\bf 0.00022 (0.00013)}  \\
    \midrule
    SGLD  & 1e-5   & 2500   &  0.00029 (0.00016) & 0.00031 (0.00016) \\
    SGLD  & 1e-5   & 5000   &  0.00021 (0.00010) & 0.00021 (0.00013) \\
    {\bf SGLD}  & \bf {1e-5}   & \bf {10000}  & {\bf 0.00019 (0.00012)} & {\bf 0.00018 (0.00010)}  \\
    \midrule
    SGHMC  & 1e-5   & 2500   &  0.00020 (0.00014) & 0.00021 (0.00021) \\
    SGHMC  & 1e-5   & 5000   &  0.00016 (0.00009) & 0.00017 (0.00011) \\
    {\bf SGHMC}  & \bf {1e-5}   & \bf {10000}  & {\bf 0.00016 (0.00010)} & {\bf 0.00016 (0.00010)}  \\
    \midrule
    DQN         & 1e-3    & -  & 0.10890 (0.14942) & 0.08630 (0.10870)  \\
    BootDQN         & 1e-3    & -  & 0.11200 (0.18036) & 0.08757 (0.14150)  \\
    QR-DQN         & 1e-2    & -  & 0.00635 (0.00453) & 0.00583 (0.00342)  \\
    \bf{KOVA}          & \bf{1}       & -  & \bf{0.00584 (0.00812)} & \bf{0.00533 (0.00771)}  \\
    
    \bottomrule
  \end{tabular}
  }
\end{table}

\begin{table}[htbp]
  \caption{Trimmed means of coverage rates and prediction interval widths 
   over 100 runs}
  \label{table:CR mean}
  \centering
  \adjustbox{max width=\textwidth}{
  \begin{tabular}{lcccccccccc}
    \toprule
    \multicolumn{3}{c}{Description}    &\multicolumn{2}{c}{East}   &\multicolumn{2}{c}{North}    \\
    \cmidrule(r){1-3}  \cmidrule(r){4-5}  \cmidrule(r){6-7} 
    Algorithm     &  $\epsilon_t$  & $\mathcal{N}$ & CR  & Range  & CR & Range  \\
    \midrule
    \bf{LKTD} & \bf {1e-5}   & \bf {2500}   & \bf{ 0.94419 (0.00946)} & \bf{0.35163 (0.03169)} & \bf{ 0.94394 (0.01004)}  & \bf{0.35271 (0.03257)}  \\
    \bf{LKTD} & \bf {1e-5}   & \bf {5000}   & \bf{ 0.94458 (0.00952)} & \bf{0.32217 (0.03195)} & \bf{ 0.94440 (0.00989)}  & \bf{0.32079 (0.03223)}  \\
    \bf{LKTD} & \bf {1e-5}   & \bf {10000}  & \bf{ 0.94577 (0.01003)} & \bf{0.31259 (0.03316)} & \bf{ 0.94537 (0.00979)}  & \bf{0.31198 (0.03327)}  \\
    
    \midrule
    \bf{SGLD} & \bf {1e-5}   & \bf {2500}   & \bf{ 0.94411 (0.01316)} & \bf{0.38984 (0.02707)} & \bf{ 0.94474 (0.01141)}  & \bf{0.38938 (0.02712)}  \\
    \bf{SGLD} & \bf {1e-5}   & \bf {5000}   & \bf{ 0.94589 (0.01191)} & \bf{0.35993 (0.02893)} & \bf{ 0.94618 (0.00924)}  & \bf{0.35930 (0.02813)}  \\
    \bf{SGLD} & \bf {1e-5}   & \bf {10000}  & \bf{ 0.94636 (0.00878)} & \bf{0.33496 (0.02148)} & \bf{ 0.94648 (0.00906)}  & \bf{0.33559 (0.02167)}  \\
    
    \midrule
    \bf{SGHMC} & \bf {1e-5}   & \bf {2500}   & \bf{ 0.94633 (0.00890)} & \bf{0.34193 (0.02125)} & \bf{ 0.94578 (0.00931)}  & \bf{0.34144 (0.02140)}  \\
    \bf{SGHMC} & \bf {1e-5}   & \bf {5000}   & \bf{ 0.94704 (0.00854)} & \bf{0.33553 (0.02439)} & \bf{ 0.94622 (0.00885)}  & \bf{0.33586 (0.02458)}  \\
    \bf{SGHMC} & \bf {1e-5}   & \bf {10000}  & \bf{ 0.94682 (0.00893)} & \bf{0.32756 (0.02000)} & \bf{ 0.94659 (0.00874)}  & \bf{0.32659 (0.01990)}  \\
    
\midrule
    DQN         & 1e-3    & -  & 0.41132 (0.20317) & 0.23791 (0.05742) & 0.41142 (0.19289) & 0.23736 (0.06317) \\
    BootDQN     & 1e-3    & -  & 0.37995 (0.18066) & 0.19207 (0.04146) & 0.39634 (0.19053) & 0.18339 (0.02263) \\
    \bf{QR-DQN}      & \bf{1e-2}    & -  & \bf{0.85690 (0.07660)} & \bf{0.40063 (0.05326)} & \bf{0.86395 (0.06097)} & \bf{0.39800 (0.05206)} \\
    KOVA        & 1       & -  & 0.24133 (0.15194) & 0.04756 (0.02534) & 0.25709 (0.15602) & 0.04987 (0.02886) \\
    \bottomrule
  \end{tabular}
  }
\end{table}

From a computational aspect, the LKTD and SGLD algorithms stand out for their efficiency and scalability, compared to the existing tracking algorithm KOVA. As detailed in Table \ref{table:computation cost}, we have recorded the average computation time required by each algorithm to execute a single parameter update, utilizing an 4-core AMD Epyc 7662 Rome processor. The findings indicate that both LKTD and SGLD scale effectively in relation to network and batch size. Their time complexities align closely with that of DQN. Conversely, the KOVA algorithm, due to its reliance on the calculation of the Jacobian matrix and matrix inversion, proves to be computationally less efficient.

\begin{table}[htbp]
  \caption{Computation time for the indoor escaping example}
  \label{table:computation cost}
  \centering
  \adjustbox{max width=\textwidth}{
  \begin{tabular}{lccccccccrr}
    \toprule        
    Algorithm  & hidden layer & batch size & gradient steps (iterations) & cpu time $(\times 10^{-3})$  & time per iteration\\
    \midrule
    LKTD  & [32, 32]  &  100 & 5 &   6.63 &  1.326 \\
    LKTD  & [32, 32]  &  200 & 5 &   7.36 &  1.472 \\
    LKTD  & [64, 64]  &  100 & 5 &   7.43 &  1.486 \\
    \midrule
    
    SGLD  & [32, 32]  &  100 & 5 &   7.15 &  1.430  \\
    SGLD  & [32, 32]  &  200 & 5 &   7.44 &  1.488  \\
    SGLD  & [64, 64]  &  100 & 5 &   7.36 &  1.472  \\
    \midrule
    
    SGHMC  & [32, 32] &  100 & 5 &   7.47 &  1.494  \\
    SGHMC  & [32, 32] &  200 & 5 &   8.25 &  1.650  \\
    SGHMC  & [64, 64] &  100 & 5 &   8.08 &  1.616  \\
    \midrule
    
    DQN  & [32, 32]  &  100 & 1 &   1.80  &   1.80\\    
    DQN  & [32, 32]  &  200 & 1 &   2.32  &   2.32\\    
    DQN  & [64, 64]  &  100 & 1 &   1.86  &   1.86\\    
    \midrule

    BootDQN  & [32, 32]  &  100 & 1 &   2.29 &   2.29 \\    
    BootDQN  & [32, 32]  &  200 & 1 &   2.68 &   2.68 \\    
    BootDQN  & [64, 64]  &  100 & 1 &   2.26 &   2.26 \\    
    \midrule

    QR-DQN  & [32, 32]  &  100 & 1 &   2.41 &   2.41 \\    
    QR-DQN  & [32, 32]  &  200 & 1 &   2.95 &   2.95  \\    
    QR-DQN  & [64, 64]  &  100 & 1 &   2.51 &   2.51 \\    
    \midrule
    
    KOVA   & [32, 32]  &  100  & 1 &   44.20 &   44.20 \\
    KOVA   & [32, 32]  &  200  & 1 &   87.00 &   87.00 \\
    KOVA   & [64, 64]  &  100  & 1 &   251.00 &   251.00 \\

    \bottomrule
  \end{tabular}
  }
\end{table}

\subsection{Classic control problems}\label{sec: hyperparameter}

\begin{figure}
     \centering
        \begin{subfigure}[b]{1\textwidth}
         \centering
         \caption{Acrobot-v1}
         \includegraphics[width=0.9\textwidth]{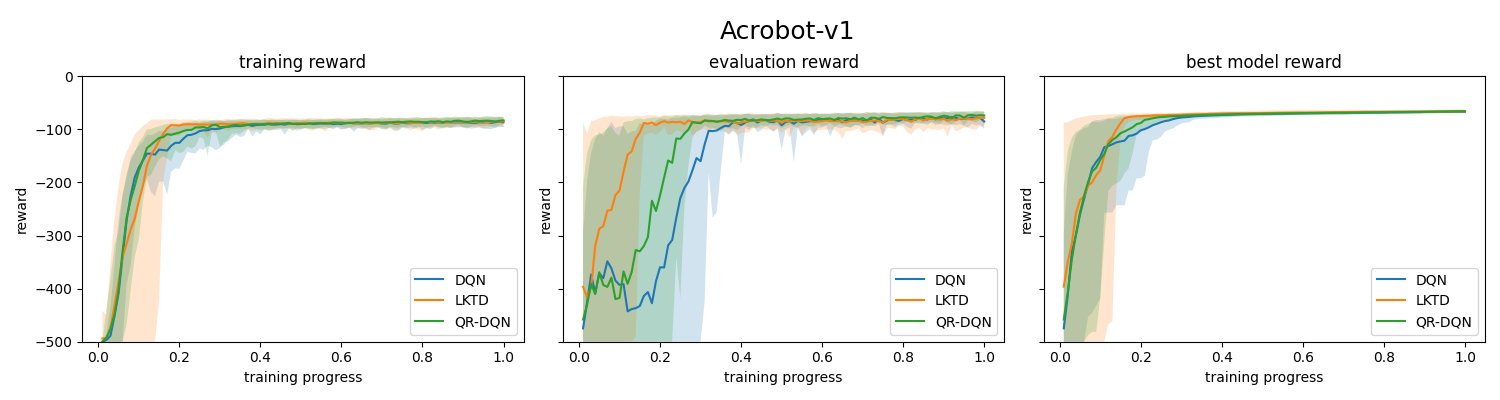}
     \end{subfigure}
     \begin{subfigure}[b]{1\textwidth}
         \centering
         \caption{LunarLander-v2}
         \includegraphics[width=0.9\textwidth]{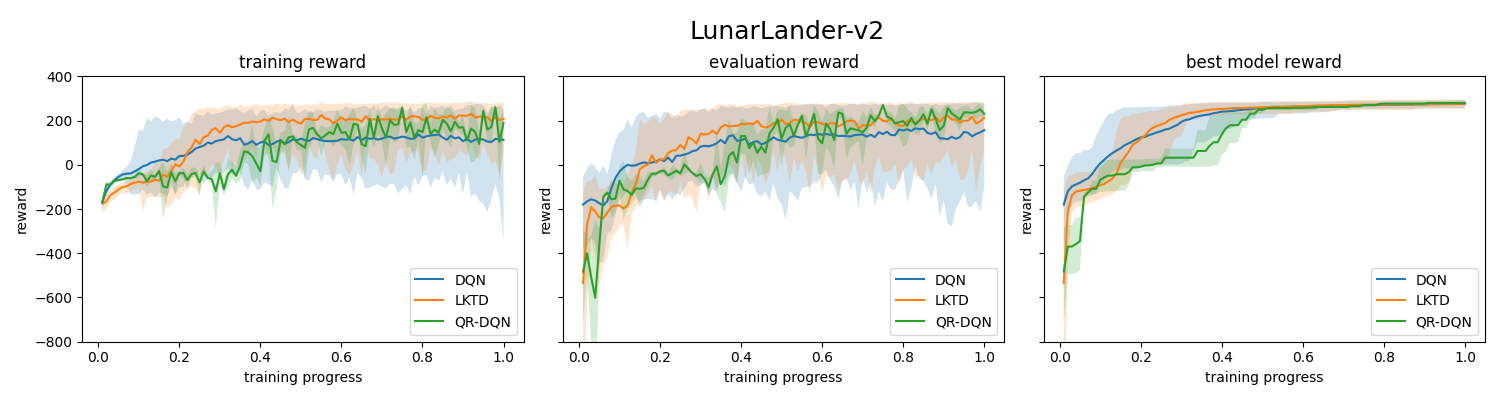}
     \end{subfigure}
     \begin{subfigure}[b]{1\textwidth}
         \centering
         \caption{MountainCar-v0}
         \includegraphics[width=0.9\textwidth]{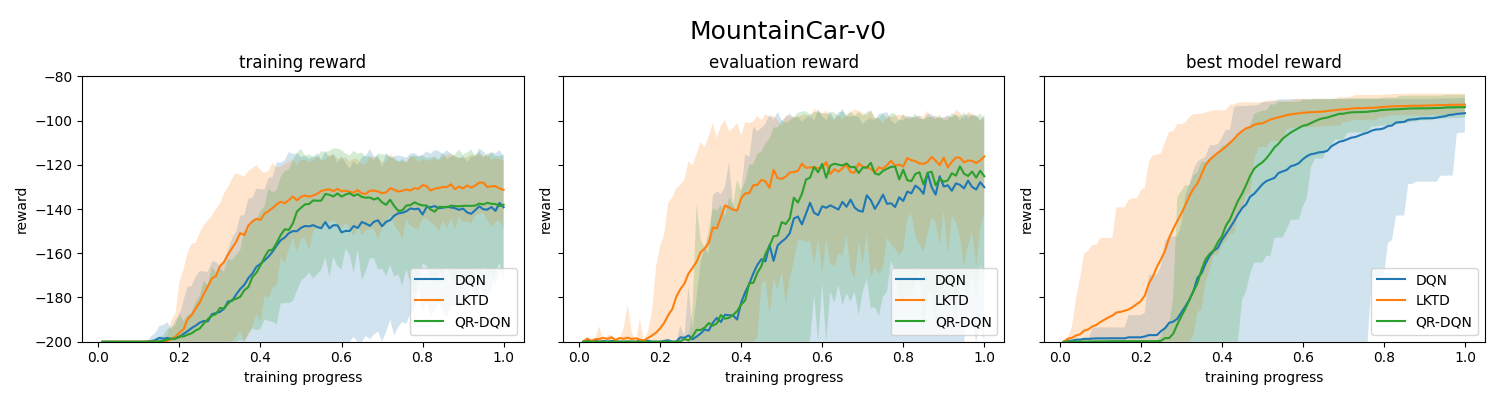}
     \end{subfigure}
        \caption{The first column shows the cumulative rewards obtained during the training process, the second column shows the testing performance without random exploration, and
        the third column shows the performance of best model learned up to the point $t$.}
        \label{fig:gym}
\end{figure}

This section evaluates the performance of LKTD on four classical control problems in OpenAI gym \citep{brockman2016openai}, including Acrobot-v1, CartPole-v1, LunarLander-v2 and MountainCar-v0. We compare LKTD with DQN and QR-DQN under the framework of RL Baselines3 Zoo \citep{rl-zoo3}. The detailed hyperparameter setting is listed in Table \ref{table:hyperparameters} and Table \ref{table:hyperparameters conti}. Each experiment is duplicated 100 times, and the training progress is recorded in Figure \ref{fig:cartpole} and Figure \ref{fig:gym}. At each time step, the best and the worst 5\% of the rewards are considered as outliers and excluded in the plots. Due to the adaptability of our sampling framework, LKTD can be easily applied to DQN algorithm by modifying the state-space model in equation (\ref{statespaceeq2}) as 
\begin{equation}
\label{eq:linear state-space model 2}
    \begin{split}
        \theta_t &= \theta_{t-1} +  \frac{\epsilon_t}{2} \nabla \log\pi(\theta_{t-1}) + w_t,\\
        \vy_t &= h(\bx_t,\theta_t) + \eta_t,
    \end{split}
\end{equation}
where  $h(\bx_t,\theta_t) = [Q_{\theta_t}(s_{t,1},a_{t,1}),\dots, Q_{\theta_t}(s_{t,n},a_{t,n})]$ 
and $\vy_t = \vr_t + \gamma Q_{\theta_t}(\bs_{t+1},\ba_{t+1})$.
With suitable constraints on the semi-gradient, we can modify 
Theorem \ref{thm:SGLD} to guarantee the convergence.
In the four classic control problems, LKTD shows its strength in efficient exploration and robustness.
In Figure \ref{fig:gym}, the lines represent the mean reward curves. For each algorithm, the colored area covers 90\% of the reward curves. We consider 3 types of reward measurements, training reward, evaluation reward and the best model reward. Training reward records the cumulative reward during training, which include the $\epsilon$-exploration errors. Evaluation reward calculates the mean reward over 5 testing trails at 100 time point throughout the training progress. The best evaluation reward records the performance of the best learned model.

\begin{table}[htbp]
  \caption{Hyperparameters}
  \label{table:hyperparameters}
  \centering
  \adjustbox{max width=\textwidth}{
  \begin{tabular}{lcccccccccccccc}
    \toprule
    Environment    &\multicolumn{3}{c}{CartPole-v1}   &\multicolumn{3}{c}{MountainCar-v0}                 \\
    \cmidrule(r){1-1} \cmidrule(r){2-4} \cmidrule(r){5-7} 
    Hyperparameters  & LKTD  & DQN & QR-DQN & LKTD  & DQN & QR-DQN\\
    \midrule
    learning rate  & 2.5e-5 & 2.3e-3 & 2.3e-3 & 1.0e-4 & 4.0e-3 & 4.0e-3 \\
    $\mathcal{N}$ (pseudo population) & 20000 & - & - & 20000 & - & - \\
    $\sigma_\theta$ (prior)  & 1 & - & - & 0.5 & - & - \\
    $\sigma$ (observation) & 1 & - & - & 1 & - & - \\
    target update interval     & 1 & 10 & 10 & 100 & 600 & 600\\
    $\gamma $(discount factor) & 0.99 & 0.99 & 0.99 & 0.98 & 0.98 & 0.98 \\
    training steps & 1e5 & 1e5 & 1e5 & 2e5 & 2e5 & 2e5 \\
    batch size    & 64 & 64 & 64 & 128 & 128 & 128 \\
    buffer size & 1e4 & 1e5 & 1e5 & 1e4 & 1e5 & 1e5  \\
    learning starts    & 1000 & 1000 & 1000 & 0  & 0 & 0 \\
    train freq    & 4 & 256 & 256 &  32 & 16 & 16  \\
    gradient steps    & 1 & 128 & 128 & 16 & 8 & 8 \\
    exploration fraction    & 0.16 & 0.16 & 0.16 & 0.2 & 0.2 & 0.2    \\
    exploration final eps   & 0.04 & 0.04 & 0.04 & 0.07 & 0.07 & 0.07     \\
    \bottomrule
  \end{tabular}
  }
\end{table}

\begin{table}[htbp]
  \caption{Hyperparameters (cont.)}
  \label{table:hyperparameters conti}
  \centering
  \adjustbox{max width=\textwidth}{
  \begin{tabular}{lcccccccccccccc}
    \toprule
    Environment    &\multicolumn{3}{c}{LunarLander-v2}   &\multicolumn{3}{c}{Acrobot-v1}                 \\
    \cmidrule(r){1-1} \cmidrule(r){2-4} \cmidrule(r){5-7} 
    Hyperparameters  & LKTD  & DQN & QR-DQN & LKTD  & DQN & QR-DQN\\
    \midrule
    learning rate  & 5.0e-6 & 6.3e-4 & 1.5e-3 & 5.0e-5 & 6.3e-4 & 6.3e-4 \\
    $\mathcal{N}$ (pseudo population) & 20000 & - & - & 20000 & - & - \\
    $\sigma_\theta$ (prior)  & 1 & - & - & 1 & - & - \\
    $\sigma$ (observation) & 1 & - & - & 1 & - & - \\
    target update interval     & 1 & 250 & 1 & 1 & 250 & 250 \\
    $\gamma $(discount factor) & 0.99 & 0.99 & 0.995 & 0.99 & 0.99 & 0.99 \\
    training steps & 2e5 & 2e5 & 2e5 & 1e5 & 1e5 & 1e5 \\
    batch size    & 128 & 128 & 128 & 128 & 128 & 128 \\
    buffer size & 2.5e4 & 5e4 & 1e5 & 5e4 & 5e4 & 5e4  \\
    learning starts    & 0 & 0 & 10000 & 0  & 0 & 50000 \\
    train freq    & 4 & 4 & 256 &  4 & 4 & 4  \\
    gradient steps    & 4 & 4 & 256 & 4 & 4 & 4 \\
    exploration fraction    & 0.24 & 0.12 & 0.24 & 0.12 & 0.12 & 0.12    \\
    exploration final eps   & 0.05 & 0.10 & 0.18 & 0.05 & 0.1 & 0.1     \\
    \bottomrule
  \end{tabular}
  }
\end{table}


\end{document}

%% file: math_commands.tex
\usepackage{amsmath,amsfonts,bm}

\def\eqref#1{equation~\ref{#1}}

\def\1{\bm{1}}

\def\eps{{\epsilon}}

\def\vr{{\bm{r}}}

\def\vx{{\bm{x}}}
\def\vy{{\bm{y}}}
\def\vz{{\bm{z}}}

\DeclareMathAlphabet{\mathsfit}{\encodingdefault}{\sfdefault}{m}{sl}
\SetMathAlphabet{\mathsfit}{bold}{\encodingdefault}{\sfdefault}{bx}{n}

\newcommand{\E}{\mathbb{E}}



%% file: myPKG.tex
\usepackage{graphicx}
\usepackage{amsmath,amsthm,amssymb,amsfonts}
\usepackage[italicdiff]{physics}
\usepackage[T1]{fontenc}
\usepackage{wrapfig}
\usepackage{lmodern,mathrsfs}
\usepackage[inline,shortlabels]{enumitem}
\usepackage[thinc]{esdiff}
\usepackage{mathtools}
\usepackage{microtype}
\usepackage{graphicx,wrapfig}
\usepackage{booktabs}
\usepackage{comment}
\usepackage{relsize}
\usepackage{adjustbox}
\usepackage{float}
\usepackage{multirow}
\usepackage[ruled,vlined]{algorithm2e}
\usepackage{epsfig,wrapfig,graphicx,subcaption} 
\usepackage{threeparttable,multirow}
\usepackage{color}
\usepackage{adjustbox,booktabs,arydshln,multirow,url}
  {\begin{Sbox}\begin{minipage}}%
  {\end{minipage}\end{Sbox}\fbox{\TheSbox}}

%% file: new_front.tex
\newcommand{\bx}{{\boldsymbol x}} 
\newcommand{\bz}{{\boldsymbol z}}

\newcommand{\bb}{{\boldsymbol b}}

\newcommand{\ba}{{\boldsymbol a}}

\newcommand{\br}{{\boldsymbol r}}
\newcommand{\bs}{{\boldsymbol s}}

\newcommand{\mK}{{\cal K}}

\newcommand{\btheta}{{\boldsymbol \theta}}

\newtheorem{theorem}{Theorem}
\newtheorem{lemma}{Lemma}

\newtheorem{corollary}{Corollary}

\newtheorem{remark}{Remark}
\newtheorem{assumption}{Assumption}



%% file: RL_iclr24_revise.bbl
\begin{thebibliography}{26}
\providecommand{\natexlab}[1]{#1}
\providecommand{\url}[1]{\texttt{#1}}
\expandafter\ifx\csname urlstyle\endcsname\relax
  \providecommand{\doi}[1]{doi: #1}\else
  \providecommand{\doi}{doi: \begingroup \urlstyle{rm}\Url}\fi

\bibitem[Anderson et~al.(1979)Anderson, Moore, and Eslami]{EKF1979OptimalF}
Brian. D.~O. Anderson, John~B. Moore, and Mansour Eslami.
\newblock Optimal filtering.
\newblock \emph{IEEE Transactions on Systems, Man, and Cybernetics}, 12:\penalty0 235--236, 1979.

\bibitem[Bellemare et~al.(2017)Bellemare, Dabney, and Munos]{bellemare2017distributional}
Marc~G. Bellemare, Will Dabney, and Rémi Munos.
\newblock A distributional perspective on reinforcement learning, 2017.

\bibitem[Brockman et~al.(2016)Brockman, Cheung, Pettersson, Schneider, Schulman, Tang, and Zaremba]{brockman2016openai}
Greg Brockman, Vicki Cheung, Ludwig Pettersson, Jonas Schneider, John Schulman, Jie Tang, and Wojciech Zaremba.
\newblock Openai gym.
\newblock \emph{arXiv preprint arXiv:1606.01540}, 2016.

\bibitem[Chen et~al.(2015)Chen, Ding, and Carin]{chen2015convergence}
Changyou Chen, Nan Ding, and Lawrence Carin.
\newblock On the convergence of stochastic gradient mcmc algorithms with high-order integrators.
\newblock In \emph{Advances in Neural Information Processing Systems}, pp.\  2278--2286, 2015.

\bibitem[Evensen(1994)]{Evensen1994}
G.~Evensen.
\newblock Sequential data assimilation with a nonlinear quasi-geostrophic model using monte carlo methods to forecast error statistics.
\newblock \emph{Journal of Geophysical Research: Oceans}, 99\penalty0 (C5):\penalty0 10143--10162, 1994.

\bibitem[Geist \& Pietquin(2010)Geist and Pietquin]{Geist2010ACM}
Matthieu Geist and Olivier Pietquin.
\newblock Kalman temporal differences.
\newblock \emph{J. Artif. Int. Res.}, 39\penalty0 (1):\penalty0 483–532, sep 2010.
\newblock ISSN 1076-9757.

\bibitem[Jin et~al.(2018)Jin, Song, Li, Gai, Wang, and Zhang]{Jin2018RealTimeBW}
Junqi Jin, Cheng-Ning Song, Han Li, Kun Gai, Jun Wang, and Weinan Zhang.
\newblock Real-time bidding with multi-agent reinforcement learning in display advertising.
\newblock \emph{Proceedings of the 27th ACM International Conference on Information and Knowledge Management}, 2018.

\bibitem[Kalman(1960)]{Kalman1960ANA}
Rudolf~E. Kalman.
\newblock A new approach to linear filtering and prediction problems.
\newblock \emph{Journal of Basic Engineering}, 82:\penalty0 35--45, 1960.

\bibitem[Kormushev et~al.(2013)Kormushev, Calinon, and Caldwell]{Kormushev2013ReinforcementLI}
Petar Kormushev, Sylvain Calinon, and Darwin~Gordon Caldwell.
\newblock Reinforcement learning in robotics: Applications and real-world challenges.
\newblock \emph{Robotics}, 2:\penalty0 122--148, 2013.

\bibitem[Li et~al.(2016)Li, Chen, Carlson, and Carin]{Li2016PreconditionedSG}
Chunyuan Li, Changyou Chen, David~E. Carlson, and Lawrence Carin.
\newblock Preconditioned stochastic gradient langevin dynamics for deep neural networks.
\newblock In \emph{AAAI}, 2016.

\bibitem[Ma et~al.(2015)Ma, Chen, and Fox]{Ma2015ACR}
Yi-An Ma, Tianqi Chen, and Emily~B. Fox.
\newblock A complete recipe for stochastic gradient mcmc.
\newblock In \emph{NIPS}, 2015.

\bibitem[Osband et~al.(2016)Osband, Blundell, Pritzel, and Van~Roy]{BootDQN}
Ian Osband, Charles Blundell, Alexander Pritzel, and Benjamin Van~Roy.
\newblock Deep exploration via bootstrapped dqn.
\newblock In D.~Lee, M.~Sugiyama, U.~Luxburg, I.~Guyon, and R.~Garnett (eds.), \emph{Advances in Neural Information Processing Systems}, volume~29. Curran Associates, Inc., 2016.
\newblock URL \url{https://proceedings.neurips.cc/paper_files/paper/2016/file/8d8818c8e140c64c743113f563cf750f-Paper.pdf}.

\bibitem[Raffin(2020)]{rl-zoo3}
Antonin Raffin.
\newblock Rl baselines3 zoo.
\newblock \url{https://github.com/DLR-RM/rl-baselines3-zoo}, 2020.

\bibitem[Raginsky et~al.(2017)Raginsky, Rakhlin, and Telgarsky]{Raginsky2017}
Maxim Raginsky, Alexander Rakhlin, and Matus Telgarsky.
\newblock Non-convex learning via stochastic gradient {L}angevin dynamics: a nonasymptotic analysis.
\newblock In \emph{Proceedings of the 2017 Conference on Learning Theory}, pp.\  1674--1703, 2017.

\bibitem[Shashua \& Mannor(2020)Shashua and Mannor]{Shashua2020KalmanMB}
Shirli Di-Castro Shashua and Shie Mannor.
\newblock Kalman meets bellman: Improving policy evaluation through value tracking.
\newblock \emph{ArXiv}, abs/2002.07171, 2020.

\bibitem[Silver et~al.(2016)Silver, Huang, Maddison, Guez, Sifre, van~den Driessche, Schrittwieser, Antonoglou, Panneershelvam, Lanctot, Dieleman, Grewe, Nham, Kalchbrenner, Sutskever, Lillicrap, Leach, Kavukcuoglu, Graepel, and Hassabis]{Silver2016MasteringTG}
David Silver, Aja Huang, Chris~J. Maddison, Arthur Guez, L.~Sifre, George van~den Driessche, Julian Schrittwieser, Ioannis Antonoglou, Vedavyas Panneershelvam, Marc Lanctot, Sander Dieleman, Dominik Grewe, John Nham, Nal Kalchbrenner, Ilya Sutskever, Timothy~P. Lillicrap, Madeleine Leach, Koray Kavukcuoglu, Thore Graepel, and Demis Hassabis.
\newblock Mastering the game of go with deep neural networks and tree search.
\newblock \emph{Nature}, 529:\penalty0 484--489, 2016.

\bibitem[Sun et~al.(2022)Sun, Song, and Liang]{SunSLiang2022sparseDNN}
Yan Sun, Qifan Song, and Faming Liang.
\newblock Consistent sparse deep learning: Theory and computation.
\newblock \emph{\JASA}, 117:\penalty0 1981--1995, 2022.

\bibitem[Sutton(1988)]{Sutton:1988}
Richard~S. Sutton.
\newblock Learning to predict by the methods of temporal differences.
\newblock \emph{Machine Learning}, 3\penalty0 (1):\penalty0 9--44, August 1988.
\newblock URL \url{http://www.cs.ualberta.ca/~sutton/papers/sutton-88.pdf}.

\bibitem[Sutton \& Barto(2018)Sutton and Barto]{Sutton1998}
Richard~S. Sutton and Andrew~G. Barto.
\newblock \emph{Reinforcement Learning: An Introduction}.
\newblock The MIT Press, second edition, 2018.
\newblock URL \url{http://incompleteideas.net/book/the-book-2nd.html}.

\bibitem[Tripp \& Shachter(2013)Tripp and Shachter]{Tripp2013ApproximateKF}
Charles~Edison Tripp and Ross~D. Shachter.
\newblock Approximate kalman filter q-learning for continuous state-space mdps.
\newblock \emph{ArXiv}, abs/1309.6868, 2013.

\bibitem[Wan \& Van Der~Merwe(2000)Wan and Van Der~Merwe]{UKF2000}
E.A. Wan and R.~Van Der~Merwe.
\newblock The unscented kalman filter for nonlinear estimation.
\newblock In \emph{Proceedings of the IEEE 2000 Adaptive Systems for Signal Processing, Communications, and Control Symposium (Cat. No.00EX373)}, pp.\  153--158, 2000.
\newblock \doi{10.1109/ASSPCC.2000.882463}.

\bibitem[Welling \& Teh(2011)Welling and Teh]{Welling2011BayesianLV}
Max Welling and Yee~Whye Teh.
\newblock Bayesian learning via stochastic gradient langevin dynamics.
\newblock In \emph{ICML}, 2011.

\bibitem[Xu et~al.(2018{\natexlab{a}})Xu, Chen, Zou, and Gu]{XuPan2018}
Pan Xu, Jinghui Chen, Difan Zou, and Quanquan Gu.
\newblock Global convergence of langevin dynamics based algorithms for nonconvex optimization.
\newblock \emph{arXiv:1707.06618v2}, 2018{\natexlab{a}}.

\bibitem[Xu et~al.(2018{\natexlab{b}})Xu, Li, Guan, Zhang, Li, Nan, Liu, Bian, and Ye]{Xu2018LargeScaleOD}
Zhe Xu, Zhixin Li, Qingwen Guan, Dingshui Zhang, Qiang Li, Junxiao Nan, Chunyang Liu, Wei Bian, and Jieping Ye.
\newblock Large-scale order dispatch in on-demand ride-hailing platforms: A learning and planning approach.
\newblock \emph{Proceedings of the 24th ACM SIGKDD International Conference on Knowledge Discovery \& Data Mining}, 2018{\natexlab{b}}.

\bibitem[Zhang et~al.(2023)Zhang, Song, and Liang]{Zhang2021LEnKF}
Peiyi Zhang, Qifan Song, and Faming Liang.
\newblock A langevinized ensemble kalman filter for large-scale static and dynamic learning.
\newblock \emph{Statistica Sinica}, 2023.
\newblock \doi{10.5705/ss.202022.0172}.

\bibitem[Zhang et~al.(2020)Zhang, Li, Zhang, Chen, and Wilson]{zhang2020cyclical}
Ruqi Zhang, Chunyuan Li, Jianyi Zhang, Changyou Chen, and Andrew~Gordon Wilson.
\newblock Cyclical stochastic gradient mcmc for bayesian deep learning, 2020.

\end{thebibliography}
